%% file: acl_latex.tex
\pdfoutput=1
\documentclass[11pt,twocolumn]{article}
\usepackage[utf8]{inputenc}  
\usepackage{acl}
\usepackage{amsmath}
\usepackage{times}
\usepackage{latexsym}
\usepackage{graphicx,xcolor}
\usepackage{url}
\usepackage{wrapfig}
\usepackage{listings}
\usepackage{colortbl}
\usepackage{float}
\usepackage{tabularx}
\usepackage{hhline}
\usepackage{hyperref}
\usepackage{multirow}
\usepackage[T1]{fontenc}
\usepackage[most]{tcolorbox}
\usepackage{pgfplots}
\pgfplotsset{compat=newest}
\usepackage{microtype}
\usepackage{inconsolata}
\usepackage{bbding}
\usepackage{pifont}
\usepackage{wasysym}
\usepackage{longtable}
\usepackage{booktabs}
\usepackage{amssymb}
\usepackage{caption}

\definecolor{codegreen}{rgb}{0,0.6,0}
\definecolor{codegray}{rgb}{0.5,0.5,0.5}
\definecolor{codepurple}{rgb}{0.58,0,0.82}
\definecolor{codeblue}{rgb}{0.25,0.5,0.75}
\definecolor{backcolour}{rgb}{0.95,0.95,0.92}
\definecolor{codeorange}{rgb}{1,0.6,0}
\definecolor{codeyellow}{rgb}{0.7,0.7,0}
\definecolor{codeteal}{rgb}{0,0.5,0.5}

\lstdefinelanguage{json}{
    basicstyle=\ttfamily\small,
    numbers=left,
    numberstyle=\tiny\color{codegray},
    stepnumber=1,
    numbersep=5pt,
    showstringspaces=false,
    breaklines=true,
    frame=single,
    backgroundcolor=\color{backcolour},
    literate=
     *{0}{{{\color{black}0}}}{1}
      {1}{{{\color{black}1}}}{1}
      {2}{{{\color{black}2}}}{1}
      {3}{{{\color{black}3}}}{1}
      {4}{{{\color{black}4}}}{1}
      {5}{{{\color{black}5}}}{1}
      {6}{{{\color{black}6}}}{1}
      {7}{{{\color{black}7}}}{1}
      {8}{{{\color{black}8}}}{1}
      {9}{{{\color{black}9}}}{1}
      {:}{{{\color{codeblue}:}}}{1}
      {,}{{{\color{black},}}}{1}
      {\{}{{{\color{codeblue}\{}}}{1}
      {\}}{{{\color{codeblue}\}}}}{1}
      {[}{{{\color{codeblue}[}}}{1}
      {]}{{{\color{codeblue}]}}}{1}
}

\lstdefinestyle{mystyle}{
    backgroundcolor=\color{backcolour},
    commentstyle=\color{codegreen},
    keywordstyle=\color{codeblue},
    numberstyle=\tiny\color{black},
    stringstyle=\color{black},
    basicstyle=\footnotesize\ttfamily,
    breakatwhitespace=false,
    breaklines=true,
    captionpos=b,
    keepspaces=true,
    numbers=left,
    numbersep=5pt,
    showspaces=false,
    showstringspaces=false,
    showtabs=false,
    tabsize=2,
    morekeywords={null,true,false},
    keywordstyle=\color{codeblue},
    morestring=[b]",
    stringstyle=\color{black},
    literate=
     *{0}{{{\color{black}0}}}{1}
      {1}{{{\color{black}1}}}{1}
      {2}{{{\color{black}2}}}{1}
      {3}{{{\color{black}3}}}{1}
      {4}{{{\color{black}4}}}{1}
      {5}{{{\color{black}5}}}{1}
      {6}{{{\color{black}6}}}{1}
      {7}{{{\color{black}7}}}{1}
      {8}{{{\color{black}8}}}{1}
      {9}{{{\color{black}9}}}{1}
      {"}{{{\color{black}"}}}{1}
      {:}{{{\color{codeblue}:}}}{1}
      {,}{{{\color{black},}}}{1}
      {\{}{{{\color{codeblue}\{}}}{1}
      {\}}{{{\color{codeblue}\}}}}{1}
      {[}{{{\color{codeblue}[}}}{1}
      {]}{{{\color{codeblue}]}}}{1}
}

\title{
Towards Artwork Explanation in Large-scale Vision Language Models}

\author{
  Kazuki Hayashi\textsuperscript{\dag}, Yusuke Sakai\textsuperscript{\dag},\\
  \textbf{Hidetaka Kamigaito}\textsuperscript{\dag}, \textbf{Katsuhiko Hayashi}\textsuperscript{\ddag}, \textbf{Taro Watanabe}\textsuperscript{\dag} \\
  \textsuperscript{\dag}Nara Institute of Science and Technology \textsuperscript{\ddag}The University of Tokyo \\
  \texttt{\{hayashi.kazuki.hl4, sakai.yusuke.sr9, kamigaito.h, taro\}@is.naist.jp} \\
  \texttt{katsuhiko-hayashi@g.ecc.u-tokyo.ac.jp}}
  
\begin{document}
\maketitle
\begin{abstract}
Large-scale Vision-Language Models (LVLMs) output text from images and instructions, demonstrating capabilities in text generation and comprehension. 
However, it has not been clarified to what extent LVLMs possess the ability to understand the knowledge necessary for explaining images, the complex relationships between various pieces of knowledge, and how they integrate these understandings into their explanations.
To address this issue, we propose a new task: the artwork explanation generation task, along with its evaluation dataset and metrics for quantitatively assessing the understanding and utilization of knowledge about artworks. 
This task is apt for image description based on the premise that LVLMs are expected to have pre-existing knowledge of artworks, which are often subjects of wide recognition and documented information.
It consists of two parts: generating explanations from images and titles of artworks, and generating explanations using only images, thus evaluating the LVLMs' language-based and vision-based knowledge.
Alongside, we release a training dataset for LVLMs to learn explanations that incorporate knowledge about artworks.
Our findings indicate that LVLMs not only struggle with integrating language and visual information but also exhibit a more pronounced limitation in acquiring knowledge from images alone\footnote{The datasets (\textbf{ExpArt}=\textbf{Exp}lain \textbf{Art}works) are available at \url{https://huggingface.co/datasets/naist-nlp/ExpArt}}.

\end{abstract}

\section{Introduction}
In the field of Vision \& Language (V\&L), Large Language Models (LLMs) \cite{touvron2023llama, vicuna2023, qwen, jiang2023mistral} have been combined with visual encoders to create Large-scale Vision-Language Models (LVLMs) \cite{li2023blip2, liu2024llavanext, Qwen-VL, ye2023mplugowl2}. 
These models have achieved success in various V\&L benchmarks \cite{li2023seedbenchbenchmarkingmultimodalllms, fu2024mmecomprehensiveevaluationbenchmark, liu2024mmbench, bai2023touchstone}.
\input{figure/fig_1}
Despite these advancements, tasks like Visual Question Answering (VQA) \cite{zhang2022clcrossvqa, yue2023mmmu}, Image Captioning \cite{Agrawal_2019,Lin2014MicrosoftCC}, and querying models about artwork-related information \cite{garcia2020AQUA, cetinic2021iconographic, bai2021explain} have primarily focused on assessing models' abilities to handle isolated pieces of knowledge.
\input{table/table_1}
These tasks, while valuable, do not fully capture the complexity of synthesizing and explaining interconnected knowledge in real-world scenarios~\cite{kawaharazuka2024realworld}, nor the difficulty of generating coherent text to explain this knowledge. 
Current evaluations often result in superficial image descriptions, lacking extensive background knowledge and interrelationships between subjects.

A pertinent example of this limitation can be observed in the context of creative support for paintings and photographs. 
As shown in Figure \ref{fig:intro}, these models must produce explanations that integrate knowledge of the artwork's theme, historical context, associated works, and artistic movement, highlighting a gap in current capabilities. 
Since this task goes beyond simply recognizing disparate knowledge, it is crucial for LVLMs to deeply understand the interrelationships of artwork knowledge to integrate them into explanations comprehensively.

To address this gap, we propose a new task and evaluation metrics designed to measure LVLMs' capability in generating comprehensive explanations about artworks. 
Our task requires LVLMs to generate explanations in response to given instructions, based on input images and titles of artworks.

We have constructed a dataset from about 10,000 English Wikipedia articles of artworks for this task and also release a training dataset to facilitate LVLMs in learning to generate explanations involving artistic knowledge. 
Furthermore, we have evaluated LVLMs currently achieving the highest performance in various V\&L benchmarks. 
The results show that while the LVLMs retain the artistic knowledge inherited from their base LLMs, they do not adequately correlate this knowledge with the provided visual information.

\section{LVLMs}
\label{sec:contents-format1}

LVLMs \cite{li2023blip2, liu2024llavanext, Qwen-VL, ye2023mplugowl2} integrate a Vision Encoder \cite{li2023blip2} trained through contrastive learning to process visual information with Large Language Models (LLMs) \cite{touvron2023llama, vicuna2023, qwen, jiang2023mistral}. 
This integration requires further training to effectively combine vision and language capabilities. 
As a result, these LVLMs significantly outperform conventional pre-trained models, even those with over ten times more parameters \cite{NEURIPS2022_960a172b, driess2023palme}.

However, it is unclear whether the knowledge from the LLM and the Vision Encoder are appropriately aligned by the additional network layers in LVLMs~\cite{chen2024image}.
Generating explanations that involve knowledge about art especially requires careful and systematic alignment and utilization of the information from both the Vision Encoder and the LLM. 
This challenge motivates us to design a new task for LVLMs.

\section{Task and Evaluation Metrics}
\label{sec:contents-format2}

\input{figure/fig2}

\subsection{Task}
\label{subsec:prop_task}

Our task requires LVLMs to generate explanations following instructions with images and titles. Examples of the instructions are shown in Table \ref{tab:prompt_template}.
As demonstrated by these examples, each instruction is categorized into three hierarchical levels, Section, Subsection, and Sub subsection, determined by the corresponding positions in Wikipedia articles (See \S\ref{sec:contents-format2}).
The proposed task addresses the following two settings with or without titles:

\paragraph{With Title}
\label{paragraph:With Title setting}
In the context of creative assistance, the title often contains the author's intent for the artwork, and it is desirable to generate explanations considering this intent. 
In this setting, both the image and its title are inputs, testing whether LVLMs can generate appropriate explanations based on both language and visual information.

\paragraph{Without Title} 
\label{paragraph:Without Title setting}
As shown in Figure \ref{fig:intro}, there are cases where a title does not exist potentially because the artwork is in the process of creation. 
This setting tests whether LVLMs can generate appropriate explanations using only visual information from images.
Additionally, analyzing the performance changes with and without titles allows us to verify the LVLMs' pure vision-based knowledge.
Furthermore, to thoroughly assess the generalization capabilities of LVLMs, we compare two cases: 1) a seen case in which images are observed during finetuning, and 2) an unseen case in which images are not observed during finetuning.

\input{table/table_11}

\subsection{Evaluation Metrics}
\label{subsec:prop_metric}

Since our task is a kind of natural language generation (NLG), we utilize popular metrics in NLG for evaluation, i.e.,  BLEU \cite{papineni-etal-2002-bleu}, ROUGE \cite{lin-2004-rouge}, and BERTScore~\cite{Zhang*2020BERTScore:}. 
To further focus on the ability to generate explanations for artworks, we propose the following three evaluation metrics\footnote{For the formulas of each metric, see Appendix \ref{sec:appendix:metric}.}:

\paragraph{Entity Coverage}
\label{entity coverage}
We evaluate how accurately the generated text includes relevant entities (see \S\ref{sec:dataset_creation}) related to the artwork mentioned in the reference description, using two evaluation settings: exact match and partial match \cite{li-etal-2022-multispanqa}.

\paragraph{Entity F1}
\label{entity F1}
We evaluate the frequency of occurrence of entities related to the artwork found in the generated and reference explanations by F1. 
Inspired by ROUGE, we consider the highest frequency of occurrence of any entities within either the generated explanation or the reference as the upper limit of occurrence frequency to calculate precision and recall.

\paragraph{Entity Cooccurrence}
This metric evaluates not only the coverage of individual entities but also how their interrelations are contextually combined to form a coherent explanation.
Specifically, it considers pairs of entities that co-occur within a sentence and its preceding and following \( n \) sentences, and measures the coverage of these pairs to assess how well the model understands and integrates relationships among relevant knowledge.
By setting \( n \) to exceed the number of sentences in the generated explanation, co-occurring entity pairs can be captured across the entire text.
To discourage long and redundant explanations, we incorporate a length penalty inspired by BLEU \cite{papineni-etal-2002-bleu}, but designed for the opposite purpose.
Unlike BLEU’s brevity penalty, which penalizes excessively short generations, our length penalty assigns a penalty to long outputs relative to the reference text.
This design encourages models to generate concise explanations while maintaining accurate and contextually integrated knowledge.

\input{table/table_2.tex}

\section{Dataset Creation}
\label{sec:dataset_creation}
The process of dataset creation, illustrated in Figure~\ref{fig:dataflow}, involved the following steps.
Detailed dataset statistics are provided in Appendix~\ref{sec:dataset_details}.

\paragraph{STEP 1:}
We collected all the artwork articles from the English Wikipedia that have an infobox (about 10,000), divided them into sections, and created descriptive texts. 
Additionally, hyperlinked texts within the articles were extracted as entities related to the artwork. 
Each descriptive text is accompanied by four pieces of information: the title, the hierarchy of sections (i.e., Section, Subsection, Sub subsection), the image, and the entities.

\paragraph{STEP 2:}
We filtered out sections that did not contribute directly to the understanding of artwork, articles without images, and texts not specific to individual art pieces or artworks to ensure the relevance and quality of the content for analysis.

\paragraph{STEP 3:}
To prevent biases that may arise due to the notoriety of the artworks included in the LVLM's training data, we shuffled the data. First, we ranked the data using six metrics: page views, number of links, number of edits, number of references, number of language versions, and article length. We then evenly split the data into test, development, and training sets at a ratio of 1:1:8 to maintain the average ranking across these sets (Table \ref{tab:dataset_numbers}). 
As described in \S\ref{sec:contents-format2}, for the Seen set, we used training images with no overlap in reference text to prevent leakage. For the Unseen set, neither images nor reference texts are from the training set.

\paragraph{STEP 4:}
The sorted data for each set were then formatted into instructions using the templates described in Section \ref{subsec:prop_task}. 
To diversify the training data, we prepared seven different templates inspired by \newcite{flan-template} for model training.

\section{Evaluation}
\label{sec:Evaluation}
\subsection{Setup}
We evaluated four models: mPLUG-Owl2 \cite{ye2023mplugowl2}, LLaVA-NeXT \cite{liu2024llavanext}, Qwen-VL-Chat \cite{Qwen-VL}, and GPT-4-Vision \cite{openai2024gpt4technicalreport}, along with an instruction-tuned version of Qwen-VL-Chat (FT), fine-tuned by our dataset with LoRA \cite{dettmers2022gptint}.\footnote{Further details for the evaluation setup and results for other models are described in Appendix \ref{sec:Appendix setting} and Appendix \ref{sec:appendix:other model result}.}
As shown in Table \ref{tab:dataset_numbers}, the data is divided based on images. In the Few-shot setting, by utilizing this data division, to prevent answer leakage in Few-shot samples, for test (Seen) evaluations, samples were selected from the test (Unseen) set, and vice versa for test (Unseen) evaluations.

\subsection{Results}
\label{sec:result}

\input{table/table2_camera}
\input{table/table_3}

\paragraph{With and Without Title}
Table \ref{tab:result-score} shows the results.
In the "With Title" setting, GPT-4-Vision achieved the highest performance in Entity Coverage and Entity F1, with Qwen-VL-Chat (FT), Qwen-VL-Chat, and LLaVA-NeXT (Yi-34B-Chat) also showing strong performance. 
Notably, Qwen-VL-Chat (FT) reached the highest precision in Entity Cooccurrence, showcasing its exceptional ability to accurately contextualize knowledge within generated text. 
This proves the superiority of our instruction-tuning dataset. 
Additionally, considering the average reference token length is 174 in the unseen setting, the significantly lower performance of LLaVA-NeXT (Yi-34B-Chat) indicates excessive token lengths may result in redundant text, which is unsuitable for generating concise explanations.
In the "Without Title" setting, Qwen-VL-Chat (FT) outperformed GPT-4-Vision across all metrics, indicating that our dataset enables accurate knowledge association and generation from visual information. 
Comparative analysis of the models' performance in scenarios with and without titles indicated a consistent drop in performance across the board.
This observation shows the challenges of generating text based solely on visual inputs. All models, including advanced ones like GPT-4-Vision, heavily depend on text-based cues.

\paragraph{LLMs vs. LVLMs}
Table \ref{tab:result-analysis} shows the results of explanation generation in the With Title setting without images for text-only LLMs\footnote{Since LLMs do not handle visual information, we evaluate them using titles as textual cues.}. 
The results indicate that GPT-4 \cite{openai2023gpt4} achieves the highest accuracy across all metrics, demonstrating strong knowledge about artworks, followed by LLaMA2 \cite{touvron2023llama}, Vicuna \cite{vicuna2023} and Yi-34B-Chat \cite{ai2024yiopenfoundationmodels} **in this setting**. 
Conversely, Qwen-Chat \cite{qwen} performs lower. 
Additionally, the comparison of Tables \ref{tab:result-score} and \ref{tab:result-analysis} reveals the extent of text-only LLM knowledge retention through integrated vision and language learning.
Knowledge about artworks is compromised in other LVLMs due to integrated learning of vision and language.
On the other hand, Qwen-VL-Chat achieves a 10\% performance boost in titled settings, signaling successful synthesis of vision and language knowledge.

\paragraph{Few-shot vs. Fine-tuning }
The results in Table \ref{tab:result-few-shot and fine-tuning} show that Fine-tuning outperforms the pure model and Few-shot settings. While Few-shot settings improve with more shots, they do not reach Fine-tuning. Considering the average token length of 174 in the reference sentences, the reduced length in Few-shot settings suggests a focus on essential terms but yields less comprehensive explanations. In contrast, Fine-tuning allows the model to learn vocabulary and the format for generating coherent explanations, leading to better performance. However, the lack of differences between Seen and Unseen settings in Fine-tuning indicates that alignment of visual and textual information requires simultaneous learning of images and descriptions.

\section{Conclusion}
We introduced a new task, artwork explanation generation, and its dataset and metrics to quantitatively evaluate the artistic knowledge comprehension and application. 
Using LVLMs, we assessed their retention and utilization of artworks knowledge from base LLMs, with or without artwork titles. Our findings indicate that while LVLMs maintain much of the artistic knowledge from their LLM counterparts, they do slightly lose some in practice. 
Furthermore, the challenges in generating text solely based on visual inputs clearly show a significant dependency on text-based cues.

\section*{Limitations}

Our research elucidates the intricacies of integrating visual and language abilities within LVLMs, yet it encounters specific limitations that define the scope of our findings. 

\paragraph{Data Source}
A principal limitation is our reliance on the diverse authorship and open editing model of Wikipedia as our data source. 
Variations in detail, writing style, and information density across entries may lead to inconsistencies in the dataset, potentially skewing model performance and affecting the universality of our conclusions. 
Additionally, we did not filter out generic entities such as "artwork" to avoid bias. 
However, more specific entity filtering may improve dataset relevance to artworks.
Moreover, relying on Wikipedia limits our dataset to well-known artworks, omitting lesser-known but culturally significant works not featured on the platform, thereby missing a broader spectrum of artistic significance.

\paragraph{Human Evaluation}
While our current study does not include human evaluation, it is crucial to assess whether the models can provide insights beyond Wikipedia and to evaluate LVLM explanations from an expert perspective for real-world applications. Another LVLM-based image explanation task, image review generation \cite{saito2024evaluating}, conducts human evaluation using non-expert annotators. Unlike their work, our task requires expert knowledge to judge the quality of generated explanations. Thus, due to the cost, expert evaluation of generated explanations across various genres is left for future work.

\paragraph{Integration of Vision and Language Representations}
Simultaneously, our study identifies a crucial limitation in the process of integrating Vision Encoders with LLMs, particularly highlighting the models' reliance on textual cues to generate text from visual inputs. \citet{kamigaito-etal-2023-table} report the same issue when predicting infoboxes, which are kinds of summaries for Wikipedia articles.
This observation underscores the difficulty of retaining language knowledge during the integration, a problem we acknowledge without offering concrete solutions. 
This gap clearly shows the pressing need for future research to not only further investigate these issues but also to develop innovative methodologies that ensure the preservation of language knowledge amidst the integration of visual and language abilities.

\paragraph{Insufficient Artwork Knowledge in LVLMs}

The limited improvement in entity coverage by LoRA indicates the difficulty of injecting artwork knowledge into LVLMs. As a solution, we can consider injecting external knowledge into LVLMs. \citet{chen2024knowledge} introduce using knowledge graphs (KGs) as a solution. However, KGs are commonly sparse and we may need to complete them by KG completion (KGC), a task to complete missing links in KGs. Traditional KGC methods \cite{10.5555/3104482.3104584,NIPS2013_1cecc7a7} are  empirically \cite{ruffinelli2020you,ali2021pykeen} and theoretically \cite{kamigaito-hayashi-2021-unified,pmlr-v162-kamigaito22a,feng2024unified} investigated in detail, and thus, these are solid whereas the pre-trained-based KGC models can outperform them \cite{wang-etal-2022-simkgc}. On the other hand, \citet{DBLP:journals/corr/abs-2311-09109} point out the leakage problem of the pre-trained-based KGC models and the actual performance of them is uncertain. Retrieval-Augmented Generation (RAG) \cite{DBLP:journals/corr/abs-2005-11401} can be another solution if LVLMs can accept lengthy input \cite{zong2024vlicl}. 

\section*{Ethical Considerations}
In our study, we meticulously curated our dataset derived from English Wikipedia. 
During the data creation phase, we individually inspected each extracted image, carefully removing those clearly unsuitable for public disclosure, ensuring no inappropriate images were included. 
Additionally, while English Wikipedia's editors actively eliminate unnecessarily offensive content to compile an encyclopedia, as outlined on their official pages regarding offensive material\footnote{\url{https://en.wikipedia.org/wiki/Wikipedia:Offensive_material}}, bias in sources, and the use of biased or opinionated sources\footnote{\url{https://en.wikipedia.org/wiki/Wikipedia:Neutral_point_of_view\#Bias_in_sources}} \footnote{\url{https://en.wikipedia.org/wiki/Wikipedia:Reliable_sources\#Biased_or_opinionated_sources}}, it is acknowledged that English Wikipedia allows the inclusion of biased information sources. 
Consequently, our dataset might also reflect the inherent biases present in the original English Wikipedia content.
Note that in this work, we used an AI assistant tool, ChatGPT, for coding support.

\section*{Acknowledgments}
This work was supported by JSPS KAKENHI Grant Numbers JP21K17801, JP23H03458.

\bibliographystyle{acl_natbib}

\bibliography{acl_latex}

\clearpage

\appendix

\section{Supplemental Results}
\label{sec:appendix:other model result}

\subsection{Detailed Evaluation of LVLMs in 'Seen' Data Settings}
\label{sec:appendix:seen unseen result}
Table \ref{tab:appendix main result include seen setting} presents the results of Large-scale Vision-Language Models including 'seen' settings, with bold type highlighting the highest score for each metric within each group. 
In this study, we assessed the generalizability of data and the precision of models fine-tuned on 'seen' and 'unseen' data during their training phase to ascertain if the fine-tuning process enhanced the models' accuracy for images encountered during training. 
Despite the images being part of the training dataset, with sections meticulously segregated to prevent data leakage, our validation revealed no significant differences in accuracy between 'seen' and 'unseen' settings. 
This finding confirms the general applicability of the data and suggests that simply viewing images, without integrating them with relevant contextual knowledge, does not inherently contribute to accuracy improvement. 
This highlights the importance of a holistic learning approach where images are paired with pertinent information to truly boost the performance of the models.

Furthermore, it is generally impractical to create datasets that combine images corresponding to the vast amounts of text data seen during the training of LLMs and to acquire these through additional integrated learning. Additionally, during the integrated learning process from LLM to LVLM, the focus is on learning pairs of individual images and their descriptions. To develop the ability to individually recognize knowledge objects and explain them based on that recognition, as well as to understand the relationships between objects and generate comprehensive explanations, it is considered necessary to use enhancement methods such as RAG and new integrated learning techniques for LVLMs.

\subsection{Extended Analysis of Additional LVLMs}
In our research, we expanded our experimental investigation beyond the models outlined in the primary section to include Blip2 \cite{li2023blip2}, mPLUG\_Owl \cite{ye2023mplugowl}, LLaVA-NeXT (Mistral) \cite{liu2024llavanext}, LLaVA-1.5 \cite{liu2023improvedllava, liu2023llava}, InstructBlip \cite{instructblip}, and Yi-6B \cite{ai2024yiopenfoundationmodels}, integrating image and language in a manner similar to the initially described models. 
Utilizing the same experimental framework as the initial tests, we conducted a thorough assessment. 
The results, as outlined in Table \ref{tab :appendix other model result}, revealed that these additional models did not exceed the accuracy levels of those featured in the main analysis (refer to Section \ref{sec:Evaluation}). 
Additionally, a comparative examination of configurations with and without titles showed a uniform decline in efficacy, emphasizing the difficulty of deriving knowledge and translating it into explanatory text generation based purely on image data.

\subsection{Detailed Performance Metrics for Base LLMs With Title Context}
Table \ref{tab:Appendix LLM result with title setting} presents the results of an evaluation involving the base LLM models of the LVLMs discussed in Tables \ref{tab:result-score} and \ref{tab :appendix other model result}.
This evaluation additionally included tests on base models such as FLAN-T5-XL\cite{chung2022scaling}, FLAN-T5-XXL, OPT\cite{zhang2022opt}, LLaMA\cite{touvron2023llama} Mistral\cite{jiang2023mistral}, and Yi-6B, which were not featured in the main analysis. 
Since Language Models (LMs) are incapable of processing image information, the evaluation was confined to the 'With Title' setting that incorporates textual information. 
Within this context, GPT-4 showcased superior performance across all tested configurations, with Mistral, Vicuna-13B, and LLaMA2 also demonstrating strong results.

Consistent with the data presented in Table \ref{tab:result-score}, the base model for LLaVA-NeXT (Yi-34B) yielded output sequences with excessively token lengths compared to its counterparts, mirroring the behavior of its LVLM version. 
This tendency for producing longer output is illustrated when compared with other models (as depicted in Figure \ref{fig:Token length} ). 
Furthermore, when examining the accuracy of the LVLMs tested in Table \ref{tab :appendix other model result} alongside the base models in relation to our task proposal, there is a discernible decline in precision across nearly all models. 
Qwen is the exception, which highlights the nuanced challenges in effectively merging image and textual data. 
This complexity stands as a pivotal challenge for the evolution of sophisticated LVLMs.

\section{Title generation}

In our task, the titles of artworks are a crucial element of knowledge related to the artworks. 
To maintain the integrity of the analysis between the settings with and without titles setting, we intentionally omitted titles from entity recognition. However, we recognized the need to understand the performance of models in generating titles of artworks based solely on visual information. 
Therefore, we conducted an additional experiment in which we presented the models with the prompt \textbf{"Please answer the title of this artwork"} along with 963 images from the "Unseen" test set and evaluated the accuracy of title generation under two settings: Exact and Partial. 
Tables \ref{tab:Appendix Main Model title generation}, \ref{tab:Appendix Sub Model title generation part1} and \ref{tab:Appendix Sub Model title generation part2} display the accuracy results of the main models and those from additional experiments, respectively.

The results showed that GPT-4-Vision achieved the highest performance with an exact match setting at 8.97\%, followed by Qwen-VL-Chat (FT) and Qwen-VL-Chat with good performances. 
Other models scored 2\% or less, highlighting the difficulty of generating titles. 
Additionally, none of the LLaVA-NeXT models were able to correctly generate a single title.

Furthermore, Table \ref{tab:annotation_stats} shows the actual artwork titles generated by the top five models with the best accuracy in the exact match setting. 
The "Rank" in the table is used to distribute the dataset evenly at the time of its creation (refer to Section \ref{sec:contents-format2}), between famous and less famous paintings, to prevent bias. 
From the table, we can infer that a higher proportion of famous artworks with higher ranks were generated, indicating that the models have a better grasp of more famous artworks.

\section{Evaluation Metrics Formulation}
\label{sec:appendix:metric}

This section elaborates on the evaluation metrics proposed in Section \ref{subsec:prop_metric} using mathematical expressions. 
An explanation consisting of \( n \) sentences generated by the model is denoted as \( G = \{g_{1},\cdots, g_{n}\} \), and a reference explanation consisting of \( m \) sentences is denoted as \( R = \{r_{1},\cdots, r_{m}\} \).
The function \( \text{Entity}(\cdot) \) is defined to extract entities contained in the input text. 
The notation \( |G| \) represents the total number of tokens in the generated explanation, and \( |R| \) represents the total number of tokens in the reference explanation.

\paragraph{Entity Coverage (EC)} is calculated as follows:
\begin{equation}
EC(G, R) = Cov(G, R)
\end{equation}
Here, \( Cov(G, R) \) is a function returning the proportion of entities in \( R \) that are covered by \( G \). 
For partial matches, the Lowest Common Subsequence (LCS) is employed to calculate the longest matching length ratio in the generated explanation relative to the length of the reference entity.

\paragraph{Entity F1 (EF$_1$)} is computed as follows:
\begin{align}
    EF_1 &= \frac{2 \times P \times R}{P + R} \\
    P &= \frac{\sum_{e_i \in Entity(G)} \text{Count}_{\text{clip}}(e_i, G, R)}{\sum_{e_j \in Entity(G)} \#(e_j, G)}\\
    R &= \frac{\sum_{e_i \in Entity(R)} \text{Count}_{\text{clip}}(e_i, G, R)}{\sum_{e_j \in Entity(R)} \#(e_j, R)},
\end{align}
where \( \#(e_j, G) \), \( \#(e_j, R) \) are functions that count the occurrences of entity \( e_j \) in \( G \) and \( R \) respectively, and \( \text{Count}_{\text{clip}}(e_i, G, R) \) returns the lesser frequency of occurrence of \( e_i \) in either \( G \) or \( R \).

\paragraph{Entity Cooccurrence (ECooc)} is calculated by applying a length penalty \( LP \) (Eq.~\ref{eq:lp}) to the co-occurrence coverage:
\begin{align}
&ECooc(G, R) \nonumber\\
=& LP(G, R) \times Cov(Co(G), Co(R)).
\end{align}
Here, \( LP(G, R) \) is defined as:
\begin{equation}
    LP(G, R) = \exp\left(-\max\left(0.0, \ \frac{|G|}{|R|} - 1\right)\right). \label{eq:lp}
\end{equation}
This penalty is inspired by BLEU's brevity penalty \cite{papineni-etal-2002-bleu}, but is designed for the opposite purpose:
it penalizes \emph{overly long} generations to discourage redundancy.
Note that \( LP(G, R)=1 \) when \( |G| \le |R| \), and it decays exponentially only when \( |G| > |R| \).
The function \( Co(\cdot) \) returns pairs of co-occurring entities within a context window comprising a sentence and its adjacent \( n \) sentences.
Sentence segmentation was performed using the NLTK sentence splitter for this purpose.

\section{Details of experimental setting}
\label{sec:Appendix setting}
\subsection{LVLM details}

\input{supplement/lvlm_detaile}

\subsection{LLM details}

\input{supplement/llm_detaile}

\subsection{Fine-tuning and Inference settings}
\label{sec:appendix:settings}
\input{table/table_10}
In this study, to ensure a fair comparison of performance across multiple models, all experiments were conducted on a single NVIDIA RTX 6000 Ada GPU, with 8-bit quantization utilized for model generation. 
However, due to resource constraints, LLaVA-NeXT (Yi-34B-Chat) model was loaded and inferred in 4-bit mode. 
To standardize the length of tokens generated across all models, the maximum token length was set to 2048. 
The same settings were applied to each model for performance comparison purposes.

\subsection{Training Datasets}

Table \ref{tab:training_datasets} lists the datasets employed to train the models addressed in this study.

\section{Details of our created dataset}
\label{sec:dataset_details}

\subsection{Dataset section distribution }
Table \ref{tab:data_taypes_frequency} provides a comprehensive breakdown of various types of sections within the dataset, along with their frequency counts. 
In designing the test set for the "seen" setting, we meticulously considered the distribution of these sections. 
Through an analysis of the frequency of each section type, we managed to evenly split the data. 
This strategic approach ensured that the test set was constructed with a balanced representation of each section type, aiming for a more equitable and thorough evaluation process.
Due to this methodology, the division of the test set into "seen" and "unseen" portions was based on the distribution of section types, rather than the number of images. 
Consequently, the number of images in the "seen" and "unseen" parts of the test set may not be equal (refer to Table \ref{tab:dataset_numbers}). 
This was a deliberate choice to prioritize a balanced representation of section types over an equal count of images, enhancing the relevance and fairness of the evaluation process.

\input{table/table_12}

\subsection{Omitted sections}

The following sections have been omitted from this document:

\begin{itemize}
    \setlength{\itemsep}{-0.5pt} 
    \item References
    \item See also
    \item External links
    \item Sources
    \item Further reading
    \item Bibliography
    \item Gallery
    \item Footnotes
    \item Notes
    \item References Sources
    \item Bibliography (In Spanish)
    \item Bibliography (In Italian)
    \item Bibliography (In German)
    \item Bibliography (In French)
    \item Images
    \item Links
    \item List
    \item Notes and references
    \item List by location
\end{itemize}

These sections were deemed unsuitable for the task of generating descriptions of artwork in this study and were therefore removed.

\subsection{Dataset Statistics}
\label{sec:dataset_statistics}

We report detailed statistics of instruction templates and hierarchical levels for the Train, Dev, and Test splits, complementing the dataset summary in the main text.
Dev and Test employ a single controlled template with Title-Included and Title-Excluded variants, while Train introduces seven linguistic templates following the same structural format.
Table~\ref{tab:dataset_stats} summarizes the resulting distributions.

\subsection{Detailed Prompt Templates}
\label{appendix:template}

As summarized in Table~\ref{tab:dataset_stats}, we employ different prompt template designs across the Train, Dev, and Test splits to support both robust training and controlled evaluation.
For training, we use seven linguistically diverse templates (Table~\ref{tab:Prompt_Train_Templates}) to reduce sensitivity to superficial prompt variations.
These templates were selected from an initial pool of 49 candidates, generated by combining seven base structures with seven verbs (\textit{e.g.}, \textit{explore}, \textit{explain}, \textit{discuss}),
based on instruction adherence and answer correctness during model selection.

In contrast, the Dev split adopts a single unified template across all hierarchy levels (Section, Subsection, Sub subsection) to suppress stylistic variation and facilitate controlled analysis (Table~\ref{tab:Prompt_Dev_Templates}).
The Test split follows the same template design as Dev (Table~\ref{tab:Prompt_Test_Templates}),
enabling evaluation on unseen artworks without introducing prompt-style distribution shifts.

\subsection{Train Dataset Example}
\label{appendix:train_dataset}

As shown in Figures \ref{tab:train_set_format_with_title} and \ref{tab:train_set_format_without_title}, we adopted the format for fine-tuning Qwen \cite{qwen} and modified the template presented in \ref{appendix:template} into the form of figures. This format was used for model training and dataset publication.

\subsection{Entity Distribution}
\label{appendix:entity distribution}

Figures \ref{fig:title_entity} and \ref{fig:nontitle_entity} present the entity distribution within our datasets. The minimal difference in data distribution between seen and unseen cases suggests that the partitioning method described in Step 3 of Section 4 is effective. 

\section{License}
In our study we created a dataset from Wikipedia articles of artworks.
The each image is available under the Creative Commons License (CC) or other licenses. 
Specific license information for each image can be found on the Wikipedia page or the image description page for that image. 
The images in this study are used under the terms of these licenses, and links to the images are provided in the datasets we publish so that users can download the images directly. 
The images themselves are not directly published.
Therefore, our data does not infringe upon the licenses.


\input{table/table_4}

\input{table/table_5}

\clearpage

\input{figure/fig3}
\input{figure/fig3.5}

\input{table/table_6}

\input{table/table_7}

\input{table/table_8}

\clearpage

\onecolumn

\captionsetup[longtable]{position=bottom}
{\scriptsize
\centering
\begin{longtable}{p{5.0cm}cccccc}
    \toprule
    Title & Rank & mPLUG-Owl & mPLUG-Owl2 & Qwen-VL-Chat & Qwen-VL-Chat(FT) & GPT-4-Vision \\ 
    \midrule
    \endfirsthead

    \multicolumn{7}{c}%
    {{\bfseries \tablename\ \thetable{} -- continued from previous page}} \\
    \midrule
    Title & Rank & mPLUG-Owl & mPLUG-Owl2 & Qwen-VL-Chat & Qwen-VL-Chat(FT) & GPT-4-Vision \\
    \midrule
    \endhead

    \midrule \multicolumn{7}{r}{{Continued on next page}} \\ 
    \endfoot

    \bottomrule
    \endlastfoot
    
    Mona Lisa & 1 & \CheckmarkBold & \CheckmarkBold & \CheckmarkBold & \CheckmarkBold & \CheckmarkBold \\
    The Great Wave off Kanagawa & 2 & \CheckmarkBold & \CheckmarkBold &  & \CheckmarkBold & \CheckmarkBold \\
    Vitruvian Man & 3 & \CheckmarkBold & \CheckmarkBold & \CheckmarkBold & \CheckmarkBold & \CheckmarkBold \\
    Winged Victory of Samothrace & 4 & \CheckmarkBold &  &  & \CheckmarkBold & \CheckmarkBold \\
    Girl with a Pearl Earring & 5 & \CheckmarkBold & \CheckmarkBold & \CheckmarkBold & \CheckmarkBold & \CheckmarkBold \\
    The Wedding at Cana & 6 & \CheckmarkBold &  & \CheckmarkBold & \CheckmarkBold & \CheckmarkBold \\
    The Anatomy Lesson of Dr. Nicolaes Tulp & 7 & \CheckmarkBold &  &  & \CheckmarkBold & \CheckmarkBold \\
    Apollo Belvedere & 9 & \CheckmarkBold & \CheckmarkBold & \CheckmarkBold &  &  \\
    Homeless Jesus & 11 &  &  & \CheckmarkBold & \CheckmarkBold & \CheckmarkBold \\
    Raphael Rooms & 12 &  &  &  &  & \CheckmarkBold \\
    Almond Blossoms & 13 & \CheckmarkBold & \CheckmarkBold &  &  & \CheckmarkBold \\
    The Death of General Wolfe & 14 & \CheckmarkBold &  & \CheckmarkBold & \CheckmarkBold & \CheckmarkBold \\
    The Persistence of Memory & 15 & \CheckmarkBold & \CheckmarkBold & \CheckmarkBold & \CheckmarkBold & \CheckmarkBold \\
    Doni Tondo & 19 &  &  &  &  & \CheckmarkBold \\
    The Turkish Bath & 20 &  &  & \CheckmarkBold &  & \CheckmarkBold \\
    Look Mickey & 26 & \CheckmarkBold & \CheckmarkBold & \CheckmarkBold & \CheckmarkBold & \CheckmarkBold \\
    The Seven Deadly Sins and the Four Last Things & 27 & \CheckmarkBold &  & \CheckmarkBold & \CheckmarkBold & \CheckmarkBold \\
    The Conspiracy of Claudius Civilis & 28 &  &  &  &  & \CheckmarkBold \\
    La Belle Ferronnière & 31 &  &  &  &  & \CheckmarkBold \\
    The Gross Clinic & 32 &  &  &  & \CheckmarkBold & \CheckmarkBold \\
    The Wedding Dance & 33 &  &  & \CheckmarkBold & \CheckmarkBold & \CheckmarkBold \\
    Sacred and Profane Love & 35 &  &  &  &  & \CheckmarkBold \\
    The Sea of Ice & 37 &  &  & \CheckmarkBold & \CheckmarkBold &  \\
    The Geographer & 41 &  &  & \CheckmarkBold &  & \CheckmarkBold \\
    Equestrian Portrait of Charles V & 45 &  &  &  & \CheckmarkBold &  \\
    The Monk by the Sea & 49 &  &  & \CheckmarkBold &  &  \\
    My Bed & 51 &  &  & \CheckmarkBold & \CheckmarkBold & \CheckmarkBold \\
    I Saw the Figure 5 in Gold & 55 &  &  &  &  & \CheckmarkBold \\
    Peace Monument & 57 &  &  &  &  & \CheckmarkBold \\
    Littlefield Fountain & 58 &  &  &  & \CheckmarkBold & \CheckmarkBold \\
    Music in the Tuileries & 59 &  &  &  &  & \CheckmarkBold \\
    The Cornfield & 60 &  &  &  & \CheckmarkBold & \CheckmarkBold \\
    Lovejoy Columns & 62 &  &  & \CheckmarkBold & \CheckmarkBold & \CheckmarkBold \\
    The Allegory of Good and Bad Government & 64 &  &  &  &  & \CheckmarkBold \\
    Sibelius Monument & 72 &  &  & \CheckmarkBold & \CheckmarkBold & \CheckmarkBold \\
    Headington Shark & 73 &  &  &  &  & \CheckmarkBold \\
    The Great Masturbator & 75 &  &  &  &  & \CheckmarkBold \\
    Self-Portrait with Thorn Necklace and Hummingbird & 81 &  &  &  & \CheckmarkBold &  \\
    Snow Storm: Steam-Boat off a Harbour's Mouth & 83 &  &  &  &  & \CheckmarkBold \\
    Bathers at Asnières & 84 &  &  &  & \CheckmarkBold & \CheckmarkBold \\
    The Bacchanal of the Andrians & 91 &  &  & \CheckmarkBold & \CheckmarkBold &  \\
    The Painter's Studio & 95 &  &  &  & \CheckmarkBold &  \\
    Carnation, Lily, Lily, Rose & 97 &  &  & \CheckmarkBold &  & \CheckmarkBold \\
    Lady Writing a Letter with her Maid & 99 &  &  &  & \CheckmarkBold & \CheckmarkBold \\
    Two Sisters (On the Terrace) & 104 &  &  & \CheckmarkBold & \CheckmarkBold & \CheckmarkBold \\
    Lion of Belfort & 112 &  &  &  &  & \CheckmarkBold \\
    Metamorphosis of Narcissus & 114 &  &  &  &  & \CheckmarkBold \\
    Lady Seated at a Virginal & 115 &  &  &  & \CheckmarkBold &  \\
    Puerta de Alcalá & 116 &  &  &  & \CheckmarkBold & \CheckmarkBold \\
    The Three Crosses & 118 &  &  & \CheckmarkBold &  &  \\
    Statue of Paddington Bear & 119 &  &  &  & \CheckmarkBold &  \\
    Our English Coasts & 139 &  &  &  &  & \CheckmarkBold \\
    Hahn/Cock & 140 &  &  &  &  & \CheckmarkBold \\
    The Wounded Deer & 144 &  &  & \CheckmarkBold &  & \CheckmarkBold \\
    The Disrobing of Christ & 148 &  &  & \CheckmarkBold & \CheckmarkBold &  \\
    Lion of Venice & 149 &  &  & \CheckmarkBold & \CheckmarkBold & \CheckmarkBold \\
    Cross in the Mountains & 153 &  &  &  &  & \CheckmarkBold \\
    Man Writing a Letter & 164 &  & \CheckmarkBold & \CheckmarkBold &  &  \\
    Dying Slave & 165 &  &  &  &  & \CheckmarkBold \\
    Nymphs and Satyr & 168 & \CheckmarkBold &  &  &  &  \\
    Tomb of Pope Alexander VII & 172 &  &  &  & \CheckmarkBold &  \\
    Greece on the Ruins of Missolonghi & 178 &  &  &  &  & \CheckmarkBold \\
    The Basket of Apples & 186 &  &  &  & \CheckmarkBold & \CheckmarkBold \\
    James Scott Memorial Fountain & 189 &  &  &  &  & \CheckmarkBold \\
    The Death of General Mercer at the Battle of Princeton, January 3, 1777 & 193 &  &  &  &  & \CheckmarkBold \\
    Madonna of the Rabbit & 200 &  &  &  & \CheckmarkBold & \CheckmarkBold \\
    Pyramid of Skulls & 209 &  &  &  &  & \CheckmarkBold \\
    Ascending and Descending & 220 &  &  &  &  & \CheckmarkBold \\
    The Madonna of Port Lligat & 221 &  &  &  & \CheckmarkBold & \CheckmarkBold \\
    Le Pont de l'Europe & 231 &  &  &  &  & \CheckmarkBold \\
    Bratatat! & 240 &  &  &  & \CheckmarkBold &  \\
    Marie Antoinette with a Rose & 247 &  &  & \CheckmarkBold & \CheckmarkBold & \CheckmarkBold \\
    The Beguiling of Merlin & 256 &  &  & \CheckmarkBold & \CheckmarkBold &  \\
    Blob Tree & 258 & \CheckmarkBold & \CheckmarkBold & \CheckmarkBold & \CheckmarkBold & \CheckmarkBold \\
    Morning in a Pine Forest & 266 &  &  &  & \CheckmarkBold & \CheckmarkBold \\
    Swann Memorial Fountain & 271 &  &  &  &  & \CheckmarkBold \\
    Equestrian Portrait of Philip IV & 272 &  &  &  & \CheckmarkBold &  \\
    Golden Guitar & 274 &  & \CheckmarkBold & \CheckmarkBold & \CheckmarkBold & \CheckmarkBold \\
    The Blind Girl & 275 &  &  &  &  & \CheckmarkBold \\
    The Lament for Icarus & 278 &  &  &  &  & \CheckmarkBold \\
    Love's Messenger & 289 &  &  &  &  & \CheckmarkBold \\
    Arrangement in Grey and Black, No. 2: Portrait of Thomas Carlyle & 304 &  &  & \CheckmarkBold &  &  \\
    The Return of the Herd & 320 &  &  &  &  & \CheckmarkBold \\
    Statue of Henry W. Grady & 327 &  &  &  &  & \CheckmarkBold \\
    Young Ladies of the Village & 333 &  &  &  &  & \CheckmarkBold \\
    Why Born Enslaved! & 355 &  &  &  &  & \CheckmarkBold \\
    Apollo Pavilion & 358 &  &  &  &  & \CheckmarkBold \\
    Looking Into My Dreams, Awilda & 371 &  &  &  &  & \CheckmarkBold \\
    Australian Farmer & 378 & \CheckmarkBold & \CheckmarkBold & \CheckmarkBold & \CheckmarkBold & \CheckmarkBold \\
    Bust of Giuseppe Mazzini & 379 &  &  &  &  & \CheckmarkBold \\
    Wind from the Sea & 399 &  &  & \CheckmarkBold & \CheckmarkBold &  \\
    Art is a Business & 415 & \CheckmarkBold & \CheckmarkBold &  &  &  \\
    Statue of George M. Cohan & 417 & \CheckmarkBold &  & \CheckmarkBold &  &  \\
    The Union of Earth and Water & 434 &  &  &  &  & \CheckmarkBold \\
    Frederick the Great Playing the Flute at Sanssouci & 440 &  &  &  &  & \CheckmarkBold \\
    Procession in St. Mark's Square & 441 &  &  &  &  & \CheckmarkBold \\
    Larry La Trobe & 443 &  &  &  &  & \CheckmarkBold \\
    From this moment despair ends and tactics begin & 460 &  &  & \CheckmarkBold & \CheckmarkBold &  \\
    Winter Landscape with Skaters & 479 &  &  &  & \CheckmarkBold &  \\
    Bust of William H. English & 489 &  & \CheckmarkBold &  &  & \CheckmarkBold \\
    Statue of Roscoe Conkling & 507 &  &  &  &  & \CheckmarkBold \\
    Still Life and Street & 531 &  &  &  &  & \CheckmarkBold \\
    Statue of William Blackstone & 536 &  &  & \CheckmarkBold &  &  \\
    Statue of Chick Hearn & 558 &  &  &  & \CheckmarkBold &  \\
    Happy Rock & 587 & \CheckmarkBold & \CheckmarkBold & \CheckmarkBold & \CheckmarkBold & \CheckmarkBold \\
    The Revells of Christendome & 608 &  &  &  & \CheckmarkBold &  \\
    Bust of Cardinal Richelieu & 629 &  &  &  &  & \CheckmarkBold \\
    Stag Hunt & 634 &  &  & \CheckmarkBold &  &  \\
    The Drover's Wife & 679 &  &  &  & \CheckmarkBold &  \\
    My Egypt & 684 &  &  &  &  & \CheckmarkBold \\
    The Viaduct at L'Estaque & 731 &  &  &  &  & \CheckmarkBold \\
    The Repast of the Lion & 733 &  &  &  &  & \CheckmarkBold \\
    Puget Sound on the Pacific Coast & 761 &  &  &  &  & \CheckmarkBold \\
    Diana and Cupid & 768 &  &  &  & \CheckmarkBold & \CheckmarkBold \\
    Portrait of Cardinal Richelieu & 778 &  &  &  & \CheckmarkBold &  \\
    Statue of Toribio Losoya & 873 &  &  &  & \CheckmarkBold &  \\
    Statue of Valentín Gómez Farías & 877 &  &  &  & \CheckmarkBold &  \\
    \caption{List of titles that were actually output by the model with exact settings.}
    \label{tab:annotation_stats} 
\end{longtable}
} 

\clearpage

\onecolumn 

\input{table/table_13}

\input{supplement/with_title}

\input{supplement/without_title}

\newpage
\input{figure/fig4}
\input{figure/fig5}

\input{supplement/mytable}

\end{document}

%% file: figure/fig_1.tex
\begin{figure}[t]
\centering
\includegraphics[width=0.98\columnwidth]{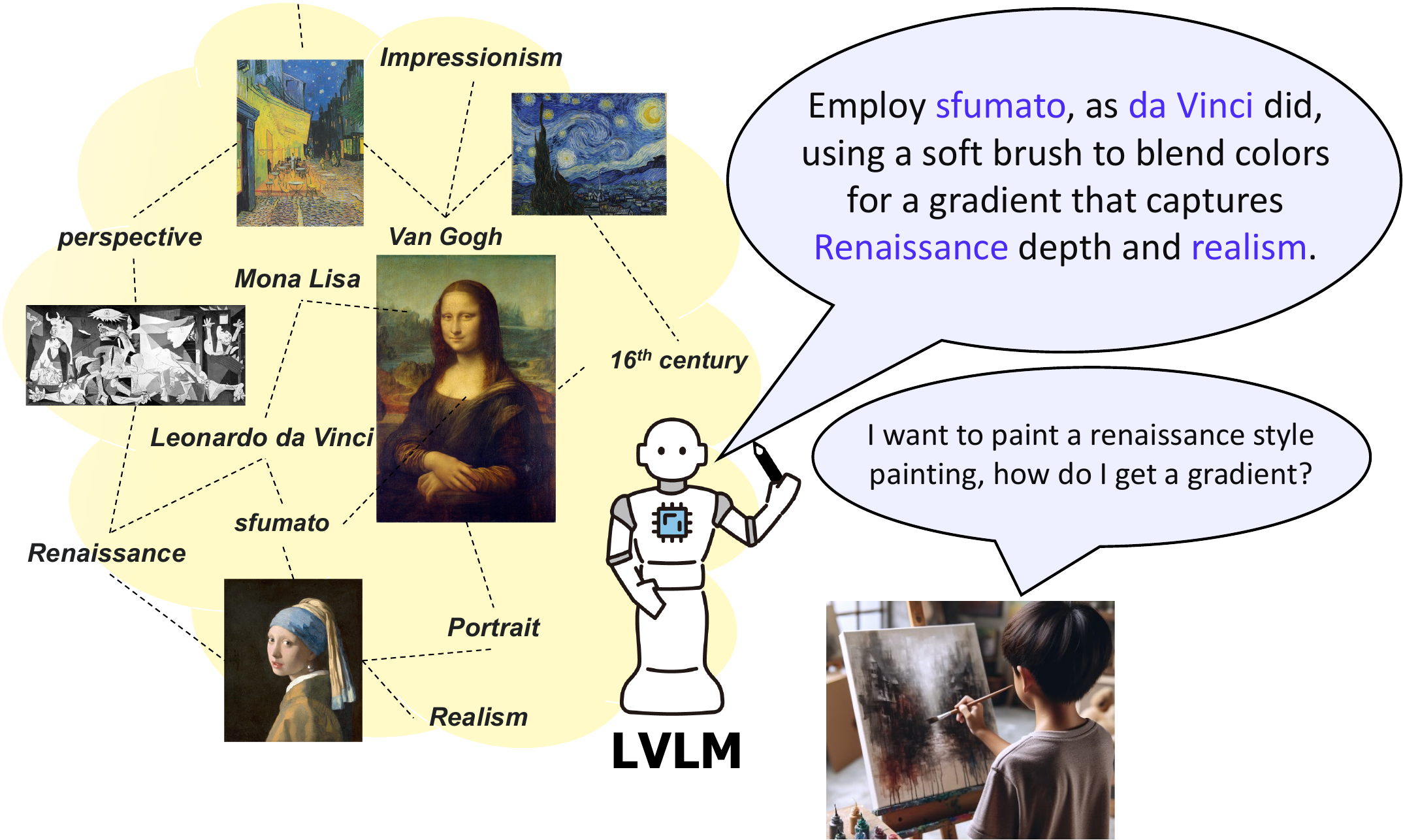}
\caption{An example of creative assistance using an LVLM, harnessing comprehensive artistic knowledge for guidance.}
\label{fig:intro}
\end{figure}

%% file: table/table_1.tex
\begin{table*}[t]
    \centering
    \resizebox{1.00\textwidth}{!}{
    \begin{tabular}{lp{6cm}p{6cm}p{6cm}}
        \toprule
        \textbf{Type} & \textbf{Template} & \textbf{Instruction} & \textbf{Output}\\
        \midrule
        Section & 
        \texttt{Explain the \textcolor{red}{\{Section\}} of this artwork, \textbf{\textcolor{blue}{\{Title\}}}.} & 
        \texttt{Explain the \textcolor{red}{History} of this artwork, \textbf{\textcolor{blue}{Mona Lisa}}.} & Of Leonardo da Vinci's works, the Mona Lisa is the only portrait whose authenticity...\\
        \midrule
        Subsection & 
        \texttt{Explain the \textcolor{red}{\{Subsection\}} about the \textcolor{red}{\{Section\}} of this artwork, \textbf{\textcolor{blue}{\{Title\}}}.} & 
        \texttt{Explain the \textcolor{red}{Creation and date} about the \textcolor{red}{History} of this artwork, \textbf{\textcolor{blue}{Mona Lisa}}.} & The record of an October 1517 visit by Louis d'Aragon states that the Mona Lisa...\\
        \midrule
        Sub subsection & 
        \texttt{Explain the \textcolor{red}{\{Sub subsection\}} about the \textcolor{red}{\{Subsection\}} of the \textcolor{red}{\{Section\}} in this artwork, \textbf{\textcolor{blue}{\{Title\}}}.} & 
        \texttt{Explain the \textcolor{red}{Creation} about the \textcolor{red}{Creation and date} of the \textcolor{red}{History} in this artwork, \textbf{\textcolor{blue}{Mona Lisa}}.} & After the French Revolution, the painting was moved to the Louvre, but spent a brief period in the bedroom of Napoleon (d. 1821) in the.... \\
        \bottomrule
    \end{tabular}
    }
    \caption{Examples of instructions for the proposed task.
Blue indicates the artwork title, and red indicates the corresponding Wikipedia sections.
Complete prompt templates are provided in Appendix~\ref{appendix:template}.}
    \label{tab:prompt_template}
\end{table*}

%% file: figure/fig2.tex
\begin{figure*}[ht]
\centering
\includegraphics[width=1.00\textwidth]{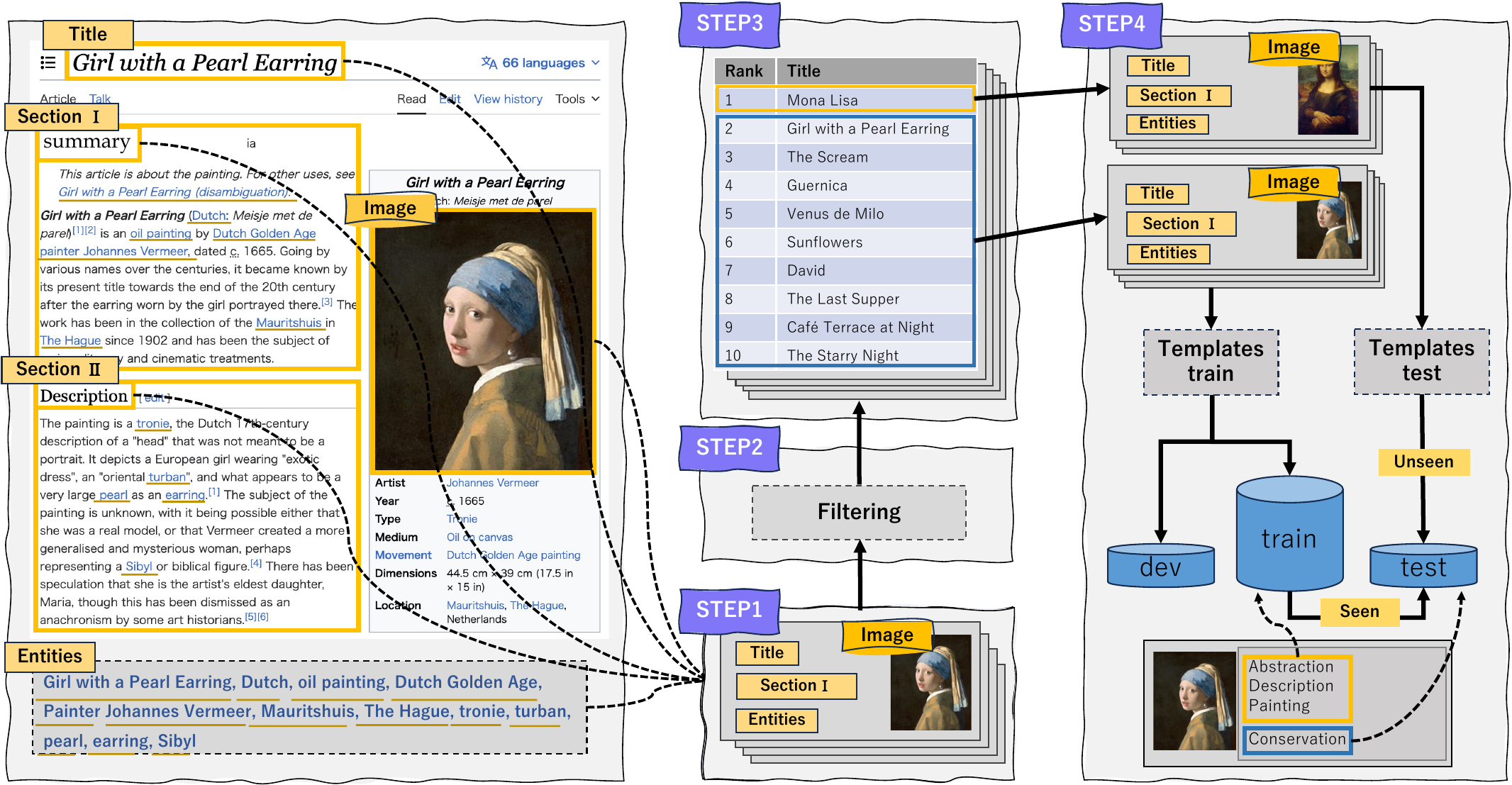}
\caption{Workflow diagram illustrating the methodology for dataset creation from Wikipedia articles on artworks, involving selection, filtering, data balancing, and instructional templating for LVLM training and evaluation.}
\label{fig:dataflow}
\end{figure*}

%% file: table/table_11.tex
\begin{table}[t]
    \centering
    \small
    \resizebox{1.00\columnwidth}{!}{\begin{tabular}{lcccc}
    \toprule
        & \textbf{Train} & \textbf{Dev} & \textbf{Test (Seen)} & \textbf{Test (Unseen)} \\ 
        \midrule
        Images & 7,704 & 963 & 2,407 & 963 \\ 
        Instruction & 18,613 & 2,677 & 2,485 & 2,597 \\
        \bottomrule
    \end{tabular}}
    \caption{Number of images and instructions in the created dataset.
Details are provided in Appendix~\ref{sec:dataset_statistics}.}
    \label{tab:dataset_numbers}
\end{table}


%% file: table/table_2.tex
\begin{table*}[t]
\centering

\resizebox{0.97\textwidth}{!}{%
\footnotesize
\begin{tabular}{@{}lccccccccccccccc@{}} 
\toprule
\multicolumn{1}{c}{\multirow{2}{*}{LVLM}} & \multirow{2}{*}{Setting} & \multirow{2}{*}{Size} & \multirow{2}{*}{BLEU} & \multicolumn{3}{c}{ROUGE} & \multirow{2}{*}{BertScore} & \multicolumn{2}{c}{Entity Cov.} & \multirow{2}{*}{Entity F1} & \multicolumn{4}{c}{Entity Cooccurrence} & \multirow{2}{*}{Avg. Length} \\
\cmidrule(lr){5-7} \cmidrule(lr){9-10} \cmidrule(lr){12-15}
& & & & 1 & 2 & L & & exact & partial & & n=0 & n=1 & n=2 & n=$\infty$ & \\
\midrule
\rowcolor[rgb]{0.956, 0.965, 0.592}
\multicolumn{16}{c}{\textbf{\textit{With Title} (Language information + Visual information)}} \\
\midrule
mPLUG-Owl2  & Unseen & 7B& 1.16 & 26.8 & 5.9 & 17.1 & 83.3 & 13.3 & 21.1 & 15.6 & 1.61 & 1.38 & 1.35 & 1.29 & 100 \\
LLaVA-NeXT (Vicuna-7B)  & Unseen & 7B & 0.81 & 16.5 & 3.7 & 11.0 & 80.8 & 9.0 & 14.1 & 10.6 & 0.83 & 0.74 & 0.73 & 0.69 & 119 \\
LLaVA-NeXT (Vicuna-13B) & Unseen & 13B & 1.18 & 17.0 & 4.1 & 10.8 & 80.5 & 11.5 & 16.4 & 13.1 & 1.12 & 1.04 & 1.02 & 0.99 & 133 \\
LLaVA-NeXT (Yi-34B) & Unseen & 34B & 0.72 & 13.9 & 3.3 & 9.5 & 80.2 & 18.5 & 27.8 & 16.1 & 0.26 & 0.22 & 0.21 & 0.19 & 869 \\
Qwen-VL-Chat & Unseen & 7B & 1.64 & 28.2 & 6.8 & 17.4 & 83.5 & 17.8 & 26.3 & 20.8 & 1.90& 1.66 & 1.63 & 1.57 & 155 \\
Qwen-VL-Chat (FT) & Unseen & 7B & \textbf{3.96} & 27.2 & \textbf{10.8} & \textbf{21.4} & \textbf{84.2} & 19.7 & 27.2 & 22.0 & \textbf{4.86} & \textbf{4.35} & \textbf{4.23} & \textbf{4.13} & 153 \\
GPT-4-Vision & Unseen & - & 2.40 & \textbf{28.6} & 7.6 & 16.3 & 83.3 & \textbf{28.4} & \textbf{37.1} & \textbf{31.6} & 3.02 & 3.00 & 2.98 & 3.05 & 264 \\
\midrule
\rowcolor[rgb]{0.6667, 0.7059, 1}
\multicolumn{16}{c}{\textbf{\textit{Without Title} (Visual information)}} \\
\midrule
mPLUG-Owl2 & Unseen & 7B & 0.21 & 23.3 & 3.58 & 15.0 & 82.3 & 4.0 & 10.5 & 4.3 & 0.26 & 0.29 & 0.26 & 0.24 & 91 \\
LLaVA-NeXT  (Vicuna-7B) & Unseen & 7B & 0.13 & 16.0 & 2.21 & 10.6 & 80.1 & 1.8 & 6.3 & 1.8 & 0.07 & 0.10 & 0.10 & 0.11 & 125 \\
LLaVA-NeXT  (Vicuna-13B) & Unseen & 13B & 0.17 & 16.6 & 2.35 & 11.0 & 80.8 & 2.1 & 7.1& 2.2 & 0.07 & 0.08 & 0.08 & 0.07 & 164 \\
LLaVA-NeXT  (Yi-34B) & Unseen & 34B & 0.15 & 11.5 & 1.88 & 8.1 & 78.7 & 3.5 & 10.5 & 2.8 & 0.03 & 0.03 & 0.02 & 0.02 & 903 \\
Qwen-VL-Chat & Unseen & 7B & 0.47 & \textbf{24.8} & 4.50 & 15.4 & 82.5 & 7.5 & 14.6& 8.4 & 0.56 & 0.60 & 0.58 & 0.55 & 128 \\
Qwen-VL-Chat (FT) & Unseen & 7B & \textbf{2.07} & 24.5 & \textbf{7.79} & \textbf{18.6} & \textbf{83.4} & \textbf{12.9} & \textbf{19.6} & \textbf{14.7} & \textbf{2.25} & \textbf{2.03} & \textbf{2.00} & \textbf{1.96} & 153 \\
GPT-4-Vision  & Unseen & - & 0.10 & 23.1 & 4.43 & 13.2 & 81.9 & 11.6 & 19.0 & 12.3 & 1.18 & 1.35 & 1.37 & 1.34 & 223 \\
\bottomrule
\end{tabular}
}
\caption{Results of LVLMs under the With Title and Without Title settings.
Bold fonts indicate the best scores for each metric.
Avg.\ Length denotes the average number of generated tokens
(see Figure~4).}
\label{tab:result-score}
\end{table*}

%% file: table/table2_camera.tex
\begin{table*}[t]
\centering
\resizebox{0.97\textwidth}{!}{%
\footnotesize
\begin{tabular}{@{}lccccccccccccccc@{}}
\toprule %
\multicolumn{1}{c}{\multirow{2}{*}{LVLM}} & \multirow{2}{*}{Setting} & \multirow{2}{*}{Size} & \multirow{2}{*}{BLEU} & \multicolumn{3}{c}{ROUGE} & \multirow{2}{*}{BertScore} & \multicolumn{2}{c}{Entity Cov.} & \multirow{2}{*}{Entity F1} & \multicolumn{4}{c}{Entity Cooccurrence} & \multirow{2}{*}{Avg. Length} \\
\cmidrule(lr){5-7} \cmidrule(lr){9-10} \cmidrule(lr){12-15}
& & & & 1 & 2 & L & & exact & partial & & n=0 & n=1 & n=2 & n=$\infty$ & \\
\midrule
\rowcolor[rgb]{0.956, 0.965, 0.592}
\multicolumn{16}{c}{\textbf{\textit{With Title} (Language information + Visual information)}} \\
\midrule
Qwen-VL-Chat & Unseen & 7B & 1.64 & \textbf{28.2} & 6.8 & 17.4 & 83.5 & 17.8 & 26.3 & 20.8 & 1.90& 1.66 & 1.63 & 1.57 & 155 \\
Qwen-VL-Chat One-shot & Unseen & 7B & 1.96 & 27.6 & 7.6 & 18.0 & 84.0 & 18.0 & 26.0 & 20.9 & 2.71 & 2.34 & 2.30 & 2.21 & 98 \\
Qwen-VL-Chat Three-shot & Unseen & 7B & 2.47 & 27.2 & 8.5 & 18.7 & \textbf{84.4} & 19.3 & \textbf{27.3} & \textbf{22.8} & 3.65 & 3.14 & 3.05 & 2.97 & 77 \\
Qwen-VL-Chat (FT) & Unseen & 7B & \textbf{3.96} & 27.2 & \textbf{10.8} & \textbf{21.4} & 84.2 & \textbf{19.7} & 27.2 & 22.0 & \textbf{4.86} & \textbf{4.35} & \textbf{4.23} & \textbf{4.13} & 153 \\
Qwen-VL-Chat (FT) One-shot & Unseen & 7B & \textbf{3.96} & 26.9 & 10.6 & 21.1 & 84.0 & \textbf{19.7} & 27.0 & 22.0 & 4.75 & 4.20 & 4.02 & 3.97 & 154 \\
Qwen-VL-Chat (FT) Three-shot & Unseen & 7B & 3.85 & 26.9 & 10.6 & 21.0 & 84.2 & 19.5 & 26.8 & 22.2 & 4.71 & 4.01 & 3.94 & 3.86 & 128 \\
\midrule
Qwen-VL-Chat & Seen & 7B & 1.69 & \textbf{27.9} & 6.7 & 17.3 & 83.4 & 16.2 & 24.5 & 19.8 & 1.87 & 1.57 & 1.54 & 1.47 & 153 \\
Qwen-VL-Chat One-shot & Seen & 7B & 2.02 & 27.3 & 7.5 & 17.8 & 84.0 & 17.4 & 25.3 & 20.8 & 2.95 & 2.49 & 2.45 & 2.36 & 95 \\
Qwen-VL-Chat Three-shot & Seen & 7B & 2.34 & 26.5 & 8.22 & 18.3 & 84.3 & 17.9 & 25.8 & 21.3 & 3.43 & 2.72 & 2.69 & 2.61 & 74 \\
Qwen-VL-Chat (FT) & Seen & 7B & \textbf{4.13} & 27.6 & \textbf{11.4} & \textbf{21.8} & 84.5 & \textbf{19.8} & \textbf{27.4} & \textbf{23.5} & \textbf{5.47} & 4.43 & 4.30 & 4.19 & 133 \\
Qwen-VL-Chat (FT) One-shot & Seen & 7B & 4.06 & 27.4 & 11.1 & 21.6 & 84.4 & \textbf{19.8} & 27.3 & 22.7 & 5.43 & \textbf{4.45} & \textbf{4.40} & \textbf{4.30} & 134 \\
Qwen-VL-Chat (FT) Three-shot & Seen & 7B & 4.05 & 27.2 & 11.1 & 21.5 & \textbf{84.6} & 19.5 & 27.0 & 22.4 & 5.22 & 4.21 & 4.19 & 4.10 & 113 \\

\midrule
\rowcolor[rgb]{0.6667, 0.7059, 1}
\multicolumn{16}{c}{\textbf{\textit{Without Title} (Visual information)}} \\
\midrule
Qwen-VL-Chat & Unseen & 7B & 0.47 & \textbf{24.8} & 4.50 & 15.4 & 82.5 & 7.5 & 14.6& 8.4 & 0.56 & 0.60 & 0.58 & 0.55 & 128 \\
Qwen-VL-Chat One-shot & Unseen & 7B & 0.65 & 23.4 & 4.81 & 15.3 & 83.0 & 8.6 & 15.4 & 9.7 & 1.15 & 1.10 & 1.04 & 1.12 & 87 \\
Qwen-VL-Chat Three-shot & Unseen & 7B & 0.69 & 22.2 & 4.95 & 15.0 & 83.3 & 9.3 & 15.6 & 10.4 & 1.21 & 1.22 & 1.17 & 1.11 & 70 \\
Qwen-VL-Chat (FT) & Unseen & 7B & \textbf{2.07} & 24.5 & \textbf{7.79} & \textbf{18.6} & 83.4 & \textbf{12.9} & \textbf{19.6} & \textbf{14.7} & 2.25 & \textbf{2.03} & \textbf{2.00} & \textbf{1.96} & 153 \\
Qwen-VL-Chat (FT) One-shot & Unseen & 7B & 1.95 & 24.1 & 7.50 & 18.3 & 83.3 & 12.6 & 19.2 & 14.3 & 2.00 & 1.92 & 1.86 & 1.84 & 152 \\
Qwen-VL-Chat (FT) Three-shot & Unseen & 7B & 2.03 & 24.3 & 7.67 & 18.4 & \textbf{83.6} & \textbf{12.9} & \textbf{19.6} & 14.6 & \textbf{2.40} & 2.00 & 1.94 & 1.91 & 131 \\
\midrule
Qwen-VL-Chat & Seen & 7B & 0.40 & 24.4 & 4.32 & 15.2 & 82.5 & 5.6 & 12.7 & 6.9 & 0.40 & 0.41 & 0.37 & 0.35 & 124 \\
Qwen-VL-Chat One-shot & Seen & 7B & 0.53 & 22.5 & 4.45 & 14.8 & 83.0 & 7.2 & 13.9 & 8.6 & 0.72 & 0.72 & 0.70 & 0.66 & 82 \\
Qwen-VL-Chat Three-shot & Seen & 7B & 0.69 & 22.2 & 4.95 & 15.0 & 83.3 & 9.3 & 15.6 & 10.4 & 1.21 & 1.22 & 1.17 & 1.11 & 68 \\
Qwen-VL-Chat (FT) & Seen & 7B & \textbf{2.09} & \textbf{24.9} & \textbf{8.00} & \textbf{18.9} & \textbf{83.8} & \textbf{12.4} & \textbf{19.4} & \textbf{15.0} & \textbf{2.19} & \textbf{1.85} & \textbf{1.82} & \textbf{1.78} & 127 \\
Qwen-VL-Chat (FT) One-shot & Seen & 7B & 1.99 & 24.4 & 7.72 & 18.5 & 83.6 & 11.5 & 18.7 & 14.0 & 1.89 & 1.55 & 1.51 & 1.48 & 130 \\
Qwen-VL-Chat (FT) Three-shot & Seen & 7B & 2.03 & 24.3 & 7.74 & 18.4 & \textbf{83.8} & 11.6 & 18.5 & 13.9 & 1.89 & 1.49 & 1.45 & 1.42 & 117 \\
\bottomrule
\end{tabular}
}
\caption{Results of Fine-tuning and Few-shot settings for LVLMs. Bold fonts indicate the best scores.}
\label{tab:result-few-shot and fine-tuning}
\end{table*}

%% file: table/table_3.tex
\begin{table}[!t]
\centering
\resizebox{0.97\columnwidth}{!}{%
\small
\setlength{\tabcolsep}{4.5pt}
\begin{tabular}{@{}lccccccccc@{}}
\toprule
\multicolumn{1}{c}{\multirow{2}{*}{LLM}} &\multicolumn{2}{c}{Entity Cov.} & \multirow{2}{*}{Entity F1} & \multicolumn{4}{c}{Entity Cooccurrence} & \multirow{2}{*}{Avg. Length}\\ 
  \cmidrule(lr){2-3} \cmidrule(lr){5-8}
  & exact & partial &  & n=0 & n=1 & n=2 & n=$\infty$ & \\
\midrule
\rowcolor[rgb]{0.956, 0.965, 0.592}
\multicolumn{9}{c}{\textbf{\textit{With Title} (Language information)}} \\
\midrule
LLaMA2 & 18.5 & 27.3 & 20.8 & 1.04 & 0.88 & 0.82 & 0.81 & 366 \\
Vicuna 7B  & 12.3 & 18.6 & 14.1 & 1.43 & 1.33 & 1.32 & 1.23 & 129 \\
Vicuna 13B & 19.4 & 28.1 & 23.0 & 2.16 & 1.99 & 1.89 & 1.77 & 209 \\
Yi-34B-Chat & 17.9 & 25.4 & 13.0 & 0.93 & 0.86 & 0.83 & 0.81 & 745 \\
Qwen-Chat & 7.6 & 11.8 & 8.5 & 0.52 & 0.43 & 0.41 & 0.40 & 106 \\
GPT-4 & \textbf{31.7} & \textbf{40.2} & \textbf{32.3} & \textbf{2.54} & \textbf{2.50} & \textbf{2.53} & \textbf{2.59} & 374 \\
\bottomrule
\end{tabular}
}
\caption{Results of LLMs (Unseen). Notations are the same as Table \ref{tab:result-score}.
}
\label{tab:result-analysis}
\end{table}

%% file: supplement/lvlm_detaile.tex
\begin{center}
\resizebox{\columnwidth}{!}{%
    \begin{tabular}{@{}lll@{}}
        \toprule
        Model                            & Base Model    & HuggingFace Name/OpenAI API                \\
        \midrule
        BLIP2 (OPT)                       & OPT             & Salesforce/blip2-opt-6.7b      \\
        BLIP2 (FLAN-T5-XL)                       & FLAN-T5-XL             & Salesforce/blip2-flan-t5-xl       \\
        BLIP2 (FLAN-T5-XXL)                      & FLAN-T5-XXL             & Salesforce/blip2-flan-t5-xxl       \\
        InstructBLIP (FLAN-T5-XL)                        & FLAN-T5-XL            & Salesforce/instructblip-flan-t5-xl \\
        InstructBLIP (FLAN-T5-XXL)                      & FLAN-T5-XXL             & Salesforce/instructblip-flan-t5-xxl        \\
        InstructBLIP (Vicuna-7B)                          & Vicuna-7B       & Salesforce/instructblip-vicuna-7b                  \\
        InstructBLIP (Vicuna-13B)                         & Vicuna-13B      & Salesforce/instructblip-vicuna-13b             \\
        Yi-VL-6B                        & Yi-6B-Chat             & 01-ai/Yi-VL-6B     \\
        mPLUG-Owl                         & LLaMA    & MAGAer13/mplug-owl-llama-7b       \\
        mPLUG-Owl2                       & LLaMA2-7B    & MAGAer13/mplug-owl2-llama2-7b     \\
        LLaVA-1.5                      & Vicuna-13B   & liuhaotian/llava-v1.5-13b    \\
        LLaVA-NeXT (Vicuna-7B)                & Vicuna-7B   & liuhaotian/llava-v1.6-vicuna-7b          \\
        LLaVA-NeXT (Vicuna-13B)           & Vicuna-13B  & liuhaotian/llava-v1.6-vicuna-13b        \\
        LLaVA-Next (Mistral)                      & Mistral       & liuhaotian/llava-v1.6-mistral-7b               \\
        LLaVA-NeXT (Yi-34B)                     & Yi-34B      & liuhaotian/llava-v1.6-34b              \\
        Qwen-VL-Chat                    & Qwen      & Qwen/Qwen-VL-Chat             \\
        GPT-4-Vision                   & -    &   gpt-4-1106-vision-preview          \\
        \bottomrule
    \end{tabular}%
}
\end{center}

%% file: supplement/llm_detaile.tex
\begin{center}
\resizebox{\columnwidth}{!}{%
    \begin{tabular}{@{}ll@{}}
        \toprule
        Model                            & HuggingFace Name                \\
        \midrule
        FLAN-T5-XL                       & google/flan-t5-xl      \\
        FLAN-T5-XXL                              & google/flan-t5-xxl       \\
        OPT                              & facebook/opt-6.7b \\
        LLaMA                          & openlm-research/open\_llama\_7b        \\
        LLaMA2                          & meta-llama/Llama-2-7b                 \\
        Mistral                     & mistralai/Mistral-7B-Instruct-v0.2             \\
        Vicuna-7B                             & lmsys/vicuna-7b-v1.5    \\
        Vicuna-13B                  & lmsys/vicuna-13b-v1.5       \\
        Qwen-Chat                      & Qwen/Qwen-7B-Chat     \\
        Yi-6B                     &  01-ai/Yi-6B   \\
        Yi-34B               & 01-ai/Yi-34B         \\    
        GPT-4              &       gpt-4-1106-preview   \\
        \bottomrule
    \end{tabular}%
}
\end{center}

%% file: table/table_10.tex
\begin{table}[ht]
\centering
\footnotesize
\begin{tabular}{lc}
\toprule
Hyper Parameter & Value \\
\midrule

torch\_dtype & bfloat16 \\
seed & 42 \\
max length & 2048 \\
warmup ratio & 0.01 \\
learning rate & 1e-5 \\
batch size & 4 \\
epoch &1 \\
lora r & 64 \\
lora alpha & 16 \\
lora dropout & 0.05 \\
lora target modules & c\_attn, attn.c\_proj, w1, w2 \\
\bottomrule
\end{tabular}
\caption{The hyper-parameters used in the experiment, and others, were set to default settings. The implementation used Transformers~\cite{wolf-etal-2020-transformers} and bitsandbytes~\cite{dettmers2022llmint8}.}
\label{tab:experiment-settings}
\end{table}

%% file: table/table_12.tex
\begin{table}[ht]
\centering
{\small
\begin{tabular}{@{}lr@{}}
\toprule
\textbf{Type} & \textbf{Frequency} \\
\midrule

Abstract & 9632 \\
Description & 2747 \\
History & 1869 \\
Background & 666 \\
Provenance & 517 \\
Reception & 346 \\
Description History & 341 \\
Analysis & 337 \\
Painting & 218 \\
Artist & 189 \\
Historical Information & 187 \\
Composition & 168 \\
Subject & 138 \\
Legacy & 127 \\
Exhibitions & 115 \\
Interpretation & 110 \\
Condition & 97 \\
In Popular Culture & 94 \\
Information & 84 \\
Design & 83 \\
Style & 78 \\
Influence & 68 \\
Creation & 65 \\
Description Style & 63 \\
Related Works & 63 \\
Acquisition & 60 \\
Context & 59 \\
Versions & 51 \\
Other Versions & 51 \\
Literature & 50 \\
Symbolism & 50 \\
The Painting & 50 \\
Attribution & 50 \\
Details & 46 \\
Notes References & 45 \\
Exhibition History & 41 \\
Location & 40 \\
Interpretations & 40 \\
Critical Reception & 39 \\
Historical Context & 39 \\
Iconography & 38 \\
Subject Matter & 37 \\
Influences & 37 \\
Exhibition & 37 \\
Commission & 36 \\
Overview & 34 \\
Analysis Description & 34 \\
Citations & 33 \\
Painting Materials & 32 \\
Controversy & 32 \\
Restoration & 32 \\
\bottomrule
\end{tabular}
}
\caption{Frequency count of data types in the dataset.}
\label{tab:data_taypes_frequency} %
\end{table}

%% file: table/table_4.tex
\begin{table*}[ht]
\centering
\resizebox{1.0\textwidth}{!}{
\footnotesize
\setlength{\tabcolsep}{6pt}
\begin{tabular}{@{}lccccccccccccccc@{}}
\toprule
\multicolumn{1}{c}{\multirow{2}{*}{LVLM}} & \multirow{2}{*}{Setting} & \multirow{2}{*}{Size} & \multirow{2}{*}{BLEU} & \multicolumn{3}{c}{ROUGE} & \multirow{2}{*}{BertScore} & \multicolumn{2}{c}{Entity Cov.} & \multirow{2}{*}{Entity F1} & \multicolumn{4}{c}{Entity Cooccurrence} & \multirow{2}{*}{Avg. Length} \\
\cmidrule(lr){5-7} \cmidrule(lr){9-10} \cmidrule(lr){12-15}
& & & & 1 & 2 & L & & exact & partial & & n=0 & n=1 & n=2 & n=$\infty$ & \\
\midrule
\rowcolor[rgb]{0.956, 0.965, 0.592}
\multicolumn{16}{c}{\textbf{\textit{With Title} (Language information + Visual information)}} \\

\midrule
mPLUG-Owl2  & Unseen & 7B& 1.16 & 26.8 & 5.9 & 17.1 & 83.3 & 13.3 & 21.1 & 15.6 & 1.61 & 1.38 & 1.35 & 1.29 & 100 \\
LLaVA-NeXT (Vicuna-7B)  & Unseen & 7B & 0.81 & 16.5 & 3.7 & 11.0 & 80.8 & 9.0 & 14.1 & 10.6 & 0.83 & 0.74 & 0.73 & 0.69 & 119 \\
LLaVA-NeXT (Vicuna-13B) & Unseen & 13B & 1.18 & 17.0 & 4.1 & 10.8 & 80.5 & 11.5 & 16.4 & 13.1 & 1.12 & 1.04 & 1.02 & 0.99 & 133 \\
LLaVA-NeXT (Yi-34B) & Unseen & 34B & 0.72 & 13.9 & 3.3 & 9.5 & 80.2 & 18.5 & 27.8 & 16.1 & 0.26 & 0.22 & 0.21 & 0.19 & 869 \\
Qwen-VL-Chat & Unseen & 7B & 1.64 & 28.2 & 6.8 & 17.4 & 83.5 & 17.8 & 26.3 & 20.8 & 1.90& 1.66 & 1.63 & 1.57 & 155 \\
Qwen-VL-Chat (FT) & Unseen & 7B & \textbf{3.96} & 27.2 & \textbf{10.8} & \textbf{21.4} & \textbf{84.2} & 19.7 & 27.2 & 22.0 & \textbf{4.86} & \textbf{4.35} & \textbf{4.23} & \textbf{4.13} & 153 \\
GPT-4-Vision & Unseen & - & 2.40 & \textbf{28.6} & 7.6 & 16.3 & 83.3 & \textbf{28.4} & \textbf{37.1} & \textbf{31.6} & 3.02 & 3.00 & 2.98 & 3.05 & 264 \\
\midrule
mPLUG-Owl2 & Seen & 7B & 1.14 & 26.6 & 5.9 & 17.0 & 83.3 & 12.5 &  20.3 & 15.1 & 1.54 & 1.29 & 1.24 & 1.17 & 94\\
LLaVA-NeXT (Vicuna-7B) & Seen & 7B & 0.78 & 16.5 & 3.5 & 10.6 & 80.7 & 7.9 & 13.0 & 9.4 & 0.74 & 0.66 & 0.63 & 0.59 & 114 \\
LLaVA-NeXT (Vicuna-13B) & Seen & 13B & 1.14 & 17.0 & 4.0 & 10.8 & 80.5 & 10.3 & 15.5 & 12.4 & 1.32 & 1.08 & 1.01 & 0.96 & 127\\
LLaVA-NeXT (Yi-34B) & Seen & 34B & 0.73 & 13.7 & 3.2 & 9.4 & 80.1 & 17.4 & 26.7 & 15.4 & 0.26 & 0.24 & 0.22 & 0.21 & 872 \\
Qwen-VL-Chat & Seen & 7B & 1.69 & \textbf{27.9} & 6.7 & 17.3 & 83.4 & 16.2 & 24.5 & 19.8 & 1.87 & 1.57 & 1.54 & 1.47 & 153 \\
Qwen-VL-Chat (FT) & Seen & 7B & \textbf{4.13} & 27.6 & \textbf{11.4} & \textbf{21.8} & \textbf{84.5}& \textbf{19.8} & \textbf{27.4} & \textbf{23.5} & \textbf{5.47} & \textbf{4.43} & \textbf{4.30} & \textbf{4.19} & 133 \\
GPT-4-Vision  & Seen & - & 2.32 & \textbf{28.3} & 7.4 & 16.2 & 83.2 & \textbf{26.4} & \textbf{34.9} & \textbf{29.7} & 2.82 & 2.71 & 2.67 & 2.63 & 254 \\

\midrule
\rowcolor[rgb]{0.6667, 0.7059, 1}
\multicolumn{16}{c}{\textbf{\textit{Without Title} (Visual information)}} \\
\midrule
mPLUG-Owl2 & Unseen & 7B & 0.21 & 23.3 & 3.58 & 15.0 & 82.3 & 4.0 & 10.5 & 4.3 & 0.26 & 0.29 & 0.26 & 0.24 & 91 \\
LLaVA-NeXT  (Vicuna-7B) & Unseen & 7B & 0.13 & 16.0 & 2.21 & 10.6 & 80.1 & 1.8 & 6.3 & 1.8 & 0.07 & 0.10 & 0.10 & 0.11 & 125 \\
LLaVA-NeXT  (Vicuna-13B) & Unseen & 13B & 0.17 & 16.6 & 2.35 & 11.0 & 80.8 & 2.1 & 7.1& 2.2 & 0.07 & 0.08 & 0.08 & 0.07 & 164 \\
LLaVA-NeXT  (Yi-34B) & Unseen & 34B & 0.15 & 11.5 & 1.88 & 8.1 & 78.7 & 3.5 & 10.5 & 2.8 & 0.03 & 0.03 & 0.02 & 0.02 & 903 \\
Qwen-VL-Chat & Unseen & 7B & 0.47 & \textbf{24.8} & 4.50 & 15.4 & 82.5 & 7.5 & 14.6& 8.4 & 0.56 & 0.60 & 0.58 & 0.55 & 128 \\
Qwen-VL-Chat (FT) & Unseen & 7B & \textbf{2.07} & 24.5 & \textbf{7.79} & \textbf{18.6} & \textbf{83.4} & \textbf{12.9} & \textbf{19.6} & \textbf{14.7} & \textbf{2.25} & \textbf{2.03} & \textbf{2.00} & \textbf{1.96} & 153 \\
GPT-4-Vision  & Unseen & - & 0.10 & 23.1 & 4.43 & 13.2 & 81.9 & 11.6 & 19.0 & 12.3 & 1.18 & 1.35 & 1.37 & 1.34 & 223 \\
\midrule
mPLUG-Owl2 & Seen & 7B & 0.14 & 22.6 & 3.37 & 14.6 & 82.2 &  2.9 & 9.2 & 3.2 & 0.19 & 0.14 & 0.13 & 0.12 & 86 \\
LLaVA-NeXT  (Vicuna-7B) & Seen & 7B & 0.11 & 15.4 & 1.95 & 10.2 & 80.0 & 1.0 & 5.6 & 1.2 & 0.05 & 0.04 & 0.06 & 0.06 & 123\\
LLaVA-NeXT  (Vicuna-13B) & Seen & 13B & 0.11 & 16.0 & 2.10 & 10.7 & 80.7 & 1.2 & 6.0 & 1.4 & 0.03 & 0.03 & 0.03 & 0.03& 154 \\
LLaVA-NeXT  (Yi-34B) & Seen & 34B & 0.10 & 11.1 & 1.71 & 7.9 & 78.6 & 2.1 & 9.2 & 1.9 & 0.01 & 0.01 & 0.01 & 0.01 & 909\\
Qwen-VL-Chat & Seen & 7B & 0.40 & 24.4 & 4.32 & 15.2 & 82.5 & 5.6 & 12.7 & 6.9 & 0.40 & 0.41 & 0.37 & 0.35 & 124 \\
Qwen-VL-Chat (FT) & Seen & 7B & \textbf{2.09} & \textbf{24.9} & \textbf{8.00} & \textbf{18.9} & \textbf{83.8} & \textbf{12.4} & \textbf{19.4} & \textbf{15.0} & \textbf{2.19} & \textbf{1.85} & \textbf{1.82} & \textbf{1.78} & 127 \\
GPT-4-Vision & Seen & - & 0.74 & 22.4 & 4.14 & 12.8 & 81.8 & 9.3 & 16.7 & 10.5 & 0.91 & 0.91 & 0.86 & 0.84 & 212 \\

\bottomrule
\end{tabular}
}
\caption{Results of LVLMs including 'seen' settings. Notations are the same as Table \ref{tab:result-score}.}
\label{tab:appendix main result include seen setting}
\end{table*}

%% file: table/table_5.tex
\begin{table*}[ht]
\centering
\resizebox{1.0\textwidth}{!}{
\footnotesize
\setlength{\tabcolsep}{6pt}
\begin{tabular}{@{}lccccccccccccccc@{}}
\toprule
\multicolumn{1}{c}{\multirow{2}{*}{LVLM}} & \multirow{2}{*}{Setting} & \multirow{2}{*}{Size} & \multirow{2}{*}{BLEU} & \multicolumn{3}{c}{ROUGE} & \multirow{2}{*}{BertScore} & \multicolumn{2}{c}{Entity Cov.} & \multirow{2}{*}{Entity F1} & \multicolumn{4}{c}{Entity Cooccurrence} & \multirow{2}{*}{Avg. Length} \\
\cmidrule(lr){5-7} \cmidrule(lr){9-10} \cmidrule(lr){12-15}
& & & & 1 & 2 & L & & exact & partial & & n=0 & n=1 & n=2 & n=$\infty$ & \\
\midrule
\rowcolor[rgb]{0.956, 0.965, 0.592}
\multicolumn{16}{c}{\textbf{\textit{With Title} (Language information + Visual information)}} \\
\midrule
BLIP2 (OPT) & Unseen & 6.7B& 0.00 & 0.1 & 0.0 & 0.1 & 76.4 & 0.0 & 0.0 & 0.0 & 0.00 & 0.00 & 0.00 & 0.00 & 0.01\\
BLIP2 (FLAN-T5-XL) & Unseen & 3B & 0.00 & 9.7 & 2.8 & 8.3 & 80.6 & 5.2 & 8.5 & 1.4 & 0.05 & 0.03 & 0.03 & 0.03 & 20 \\
BLIP2 (FLAN-T5-XXL) & Unseen & 11B & 0.01 & 2.8 & 0.5 & 2.6 & 76.5 & 0.7 & 2.4 & 0.5 & 0.01 & 0.00 & 0.00 & 0.00 & 21\\
mPLUG-Owl & Unseen & 7B & 0.17 & 15.0 & 2.4 & 10.1 & 81.8 & 4.3 & 8.6 & 4.7 & 0.35 & 0.38 & 0.40 & 0.37 & 12 \\
LLaVA-1.5 & Unseen & 13B & 1.61 & 20.8 & 5.2 & 13.2 & 81.5 & 13.4 & 19.4 & 15.8 & 1.56 & 1.34 & 1.33 & 1.26 & 139\\
LLaVA-NeXT  (Mistral) & Unseen & 7B & 1.32 & 24.1 & 5.7 & 15.9 & 82.4 & 12.3 & 19.6 & 14.9 & 1.44 & 1.18 & 1.15 & 1.06 & 140 \\
InstructBLIP (FLAN-T5-XL) & Unseen & 3B & 0.70 & 16.9 & 5.2 & 13.0 & 83.2 & 8.5 & 13.8 & 6.6 & 0.80 & 0.62 & 0.59 & 0.56 & 28\\
InstructBLIP (FLAN-T5-XXL) & Unseen & 11B & 1.00 & 16.4 & 4.6 & 12.0 & 81.7 & 8.6 & 13.8 & 9.3 & 1.00 & 0.75 & 0.73 & 0.71 & 54\\
InstructBLIP (Vicuna-7B) & Unseen & 7B & 1.44 & 23.5 & 6.2 & 15.7 & 83.3 & 12.6 & 19.2 & 14.2 & 1.79 & 1.50 & 1.44 & 1.38 & 58\\
InstructBLIP (Vicuna-13B) & Unseen & 13B & 1.11 & 25.9 & 6.2 & 17.2 & 83.6 & 11.8 & 18.8 & 13.7 & 1.42 & 1.19 & 1.16 & 1.09 & 50\\
Yi-VL-6B & Unseen & 6B & 1.07 & 26.2 & 5.7 & 16.6 & 82.9 & 12.9 & 20.8 & 15.1 & 1.37 & 1.24 & 1.27 & 1.21 & 147 \\
\rowcolor[rgb]{0.8275, 0.8196, 0.8314}
Qwen-VL-Chat & Unseen & 7B & 1.64 & 28.2 & 6.8 & 17.4 & 83.5 & 17.8 & 26.3 & 20.8 & 1.90& 1.66 & 1.63 & 1.57 & 155 \\
\rowcolor[rgb]{0.8275, 0.8196, 0.8314}
Qwen-VL-Chat (FT) & Unseen & 7B & \textbf{3.96} & 27.2 & \textbf{10.8} & \textbf{21.4} & \textbf{84.2} & 19.7 & 27.2 & 22.0 & \textbf{4.86} & \textbf{4.35} & \textbf{4.23} & \textbf{4.13} & 153 \\
\rowcolor[rgb]{0.8275, 0.8196, 0.8314}
GPT-4-Vision & Unseen & - & 2.40 & \textbf{28.6} & 7.6 & 16.3 & 83.3 & \textbf{28.4} & \textbf{37.1} & \textbf{31.6} & 3.02 & 3.00 & 2.98 & 3.05 & 264 \\

\midrule
BLIP2 (OPT) & Seen & 6.7B & 0.00 & 2.0 & 0.0 & 1.2 & 77.5 & 0.0 & 1.8  & 0.0 & 0.00 & 0.00 & 0.00 & 0.00 & 0.01\\
BLIP2 (FLAN-T5-XL) & Seen & 3B & 0.01 & 9.9 & 3.0 & 8.5 & 80.7 & 5.2 & 8.3 & 1.7 & 0.07 & 0.03 & 0.03 & 0.03 & 17\\
BLIP2 (FLAN-T5-XXL) & Seen & 11B & 0.01 & 2.9 & 0.5 & 2.7 & 76.5 & 0.9 & 2.6 & 0.6 & 0.04 & 0.03 & 0.03 & 0.03 & 21\\
mPLUG-Owl & Seen & 7B & 0.14 & 15.4 & 2.4 & 10.3 & 81.9 & 4.5 & 9.3 & 4.8 & 0.37 & 0.29 & 0.28 & 0.26 & 13\\
LLaVA-1.5 & Seen & 13B & 1.69 & 20.7 & 5.3 & 13.1 & 81.5 & 12.5 & 18.4 & 15.0 & 1.85 & 1.37 & 1.34 & 1.30 & 128\\
LLaVA-NeXT  (Mistral) & Seen & 7B & 1.41 & 24.1 & 5.6 & 16.0 & 82.3 & 11.6 & 19.1 & 14.4 & 1.49 & 1.16 & 1.06 & 1.01 & 145 \\
InstructBLIP (FLAN-T5-XL)  & Seen & 3B & 0.78 & 16.9 & 5.2 & 13.0 & 83.2 & 8.5 & 14.0 & 7.1 & 0.92 & 0.69 & 0.66 & 0.63 & 29 \\
InstructBLIP (FLAN-T5-XXL) & Seen & 11B & 0.10 & 16.6 & 4.7 & 12.2 & 81.8 & 8.7 & 14.1 & 9.3 & 1.11 & 0.90 & 0.87 & 0.84 & 54\\
InstructBLIP (Vicuna-7B) & Seen & 7B & 1.53 & 23.9 & 6.3 & 15.8 & 83.3 & 12.4 & 19.5 & 14.3 & 1.77 & 1.47 & 1.42 & 1.37 & 62\\
InstructBLIP (Vicuna-13B) & Seen & 13B & 1.11 & 25.5 & 6.1 & 16.9 & 83.5 & 10.2 & 17.3 & 12.5 & 1.26 & 1.08 & 1.01 & 0.97 & 51\\
Yi-VL-6B & Seen & 6B & 1.00 & 25.8 & 5.5 & 16.3 & 82.7 & 11.5 & 19.9 & 13.6 & 1.00 & 0.80 & 0.78 & 0.75 & 149\\
\rowcolor[rgb]{0.8275, 0.8196, 0.8314}
Qwen-VL-Chat & Seen & 7B & 1.69 & \textbf{27.9} & 6.7 & 17.3 & 83.4 & 16.2 & 24.5 & 19.8 & 1.87 & 1.57 & 1.54 & 1.47 & 153 \\
\rowcolor[rgb]{0.8275, 0.8196, 0.8314}
Qwen-VL-Chat (FT) & Seen & 7B & \textbf{4.13} & 27.6 & \textbf{11.4} & \textbf{21.8} & \textbf{84.5}& \textbf{19.8} & \textbf{27.4} & \textbf{23.5} & \textbf{5.47} & \textbf{4.43} & \textbf{4.30} & \textbf{4.19} & 133 \\
\rowcolor[rgb]{0.8275, 0.8196, 0.8314}
GPT-4-Vision  & Seen & - & 2.32 & \textbf{28.3} & 7.4 & 16.2 & 83.2 & \textbf{26.4} & \textbf{34.9} & \textbf{29.7} & 2.82 & 2.71 & 2.67 & 2.63 & 254 \\
\midrule
\rowcolor[rgb]{0.6667, 0.7059, 1}
\multicolumn{16}{c}{\textbf{\textit{Without Title} (Visual information)}} \\
\midrule
BLIP2 (OPT) & Unseen & 6.7B & 0.00 & 4.1 & 0.00 & 4.1 & 79.8 & 0.0 & 0.0 & 0.0 & 0.00 & 0.00 & 0.00 & 0.00& 0.01\\
BLIP2 (FLAN-T5-XL) & Unseen & 3B & 0.01 & 8.9 & 1.47 & 7.5 & 81.2 & 2.1 & 5.0 & 1.1 & 0.01 & 0.00 & 0.00 & 0.00 & 15\\
BLIP2 (FLAN-T5-XXL) & Unseen & 11B & 0.00 & 2.5 & 0.16 & 2.4 & 75.8 & 0.6 & 1.7 & 0.2 & 0.00 & 0.00 & 0.00 & 0.00 & 18\\
mPLUG-Owl & Unseen & 7B & 0.14 & 18.1 & 2.59 & 11.9 & 82.1 & 2.2 & 7.2 & 2.4 & 0.13 & 0.10 & 0.08 & 0.08 & 21\\
LLaVA-1.5 & Unseen & 13B & 0.21 & 17.8 & 2.70 & 11.7 & 81.4 & 2.7 & 7.9 & 2.6 & 0.11 & 0.15 & 0.15 & 0.15 & 158\\
LLaVA-NeXT  (Mistral) & Unseen & 7B & 0.16 & 21.1 & 2.77 & 14.1 & 81.3 & 2.3 & 8.0 & 2.3 & 0.08 & 0.11 & 0.12 & 0.12 & 132 \\
InstructBLIP (FLAN-T5-XL)  & Unseen & 3B & 0.08 & 13.0 & 2.17 & 10.0 & 82.4 & 2.7 & 6.6 & 2.3 & 0.13 & 0.07 & 0.08 & 0.07 & 28 \\
InstructBLIP (FLAN-T5-XXL) & Unseen & 11B & 0.16 & 12.5 & 2.11 & 9.3 & 81.1 &3.0 & 6.9 & 2.7 & 0.16 & 0.13 & 0.11 & 0.11 & 41 \\
InstructBLIP (Vicuna-7B) & Unseen & 7B & 0.49 & 22.9 & 4.47 & 15.2 & 82.9 & 6.4 & 12.9 & 7.1 & 0.55 & 0.58 & 0.56 & 0.49 & 83\\
InstructBLIP (Vicuna-13B) & Unseen & 13B & 0.39 & 23.5 & 4.31 & 15.8 & 82.8 & 4.8 & 11.5 & 5.2 & 0.37 & 0.33 & 0.31 & 0.28 & 85 \\
Yi-VL-6B & Unseen & 6B & 0.37 & 23.4 & 4.08 & 15.1 & 82.0 & 5.4 & 12.2 & 5.7 & 0.35 & 0.36 & 0.35 & 0.34 & 158\\
\rowcolor[rgb]{0.8275, 0.8196, 0.8314}
Qwen-VL-Chat & Unseen & 7B & 0.47 & \textbf{24.8} & 4.50 & 15.4 & 82.5 & 7.5 & 14.6& 8.4 & 0.56 & 0.60 & 0.58 & 0.55 & 128 \\
\rowcolor[rgb]{0.8275, 0.8196, 0.8314}
Qwen-VL-Chat (FT) & Unseen & 7B & \textbf{2.07} & 24.5 & \textbf{7.79} & \textbf{18.6} & \textbf{83.4} & \textbf{12.9} & \textbf{19.6} & \textbf{14.7} & \textbf{2.25} & \textbf{2.03} & \textbf{2.00} & \textbf{1.96} & 153 \\
\rowcolor[rgb]{0.8275, 0.8196, 0.8314}
GPT-4-Vision  & Unseen & - & 0.10 & 23.1 & 4.43 & 13.2 & 81.9 & 11.6 & 19.0 & 12.3 & 1.18 & 1.35 & 1.37 & 1.34 & 223 \\
\midrule
BLIP2 (OPT) & Seen & 6.7B & 0.00 & 2.3 & 0.00 & 2.3 & 78.4 &  0.0 & 2.1 & 0.0 & 0.00 & 0.00 & 0.00 & 0.00 & 0.03\\
BLIP2 (FLAN-T5-XL) & Seen & 3B & 0.00 & 9.0 & 1.50 & 7.6 & 81.4 & 1.7 & 4.5 & 1.0 & 0.01 & 0.01 & 0.01 & 0.01 & 13\\
BLIP2 (FLAN-T5-XXL) & Seen & 11B & 0.00 & 2.6 & 0.16 & 2.5 & 75.7 & 0.4 & 1.6 & 0.2 & 0.00 & 0.00 & 0.00 & 0.00 & 18\\
mPLUG-Owl & Seen & 7B & 0.08 & 18.4 & 2.64 & 12.1 & 82.1 & 1.9 & 6.9 & 2.5 & 0.08 & 0.05 & 0.04 & 0.04 & 23\\
LLaVA-1.5 & Seen & 13B & 0.13 & 17.7 & 2.55 & 11.6 & 81.3 & 1.3 & 6.4 & 1.4 & 0.07 & 0.05 & 0.05 & 0.04 & 154 \\
LLaVA-NeXT  (Mistral) & Seen & 7B & 0.08 & 20.7 & 2.50 & 13.9 & 81.3 & 1.3 & 7.0 & 1.4 & 0.04 & 0.04 & 0.04 & 0.03 & 125 \\
InstructBLIP (FLAN-T5-XL)  & Seen & 3B & 0.05 & 12.5 & 1.99 & 9.6 & 82.4 & 1.9 & 5.9 & 1.9 & 0.04 & 0.06 & 0.06 & 0.06 & 26 \\
InstructBLIP (FLAN-T5-XXL) & Seen & 11B & 0.10 & 12.3 & 1.95 & 9.1 & 81.1 & 2.3 & 6.3 & 2.2 & 0.08 & 0.08 & 0.07 & 0.07 & 37 \\
InstructBLIP (Vicuna-7B) & Seen & 7B & 0.43 & 22.7 & 4.31 & 15.1 & 83.0 & 4.9 & 11.4 & 5.8 & 0.36 & 0.30 & 0.29 & 0.27 & 82 \\
InstructBLIP (Vicuna-13B) & Seen & 13B & 0.37 & 23.3 & 4.27 & 15.7 & 82.7 & 3.3 & 10.0 & 4.0 & 0.17 & 0.16 & 0.16 & 0.15 & 85\\
Yi-VL-6B & Seen & 6B & 0.33 & 23.0 & 3.86 & 14.8 & 81.9 & 4.1 & 11.2 & 4.7 & 0.19 & 0.16 & 0.15 & 0.14 & 162\\
\rowcolor[rgb]{0.8275, 0.8196, 0.8314}
Qwen-VL-Chat & Seen & 7B & 0.40 & 24.4 & 4.32 & 15.2 & 82.5 & 5.6 & 12.7 & 6.9 & 0.40 & 0.41 & 0.37 & 0.35 & 124 \\
\rowcolor[rgb]{0.8275, 0.8196, 0.8314}
Qwen-VL-Chat (FT) & Seen & 7B & \textbf{2.09} & \textbf{24.9} & \textbf{8.00} & \textbf{18.9} & \textbf{83.8} & \textbf{12.4} & \textbf{19.4} & \textbf{15.0} & \textbf{2.19} & \textbf{1.85} & \textbf{1.82} & \textbf{1.78} & 127 \\
\rowcolor[rgb]{0.8275, 0.8196, 0.8314}
GPT-4-Vision & Seen & - & 0.74 & 22.4 & 4.14 & 12.8 & 81.8 & 9.3 & 16.7 & 10.5 & 0.91 & 0.91 & 0.86 & 0.84 & 212 \\

\bottomrule
\end{tabular}
}
\caption{Comprehensive Results of Secondary (LVLMs). This includes models not highlighted in the main findings, with the gray lines representing the three models that achieved the best performance in the main evaluation. Bold type signifies the highest scores for each metric within their respective groups.}
\label{tab :appendix other model result}
\end{table*}

%% file: figure/fig3.tex
\begin{figure*}[ht]
\centering
\includegraphics[width=\textwidth]{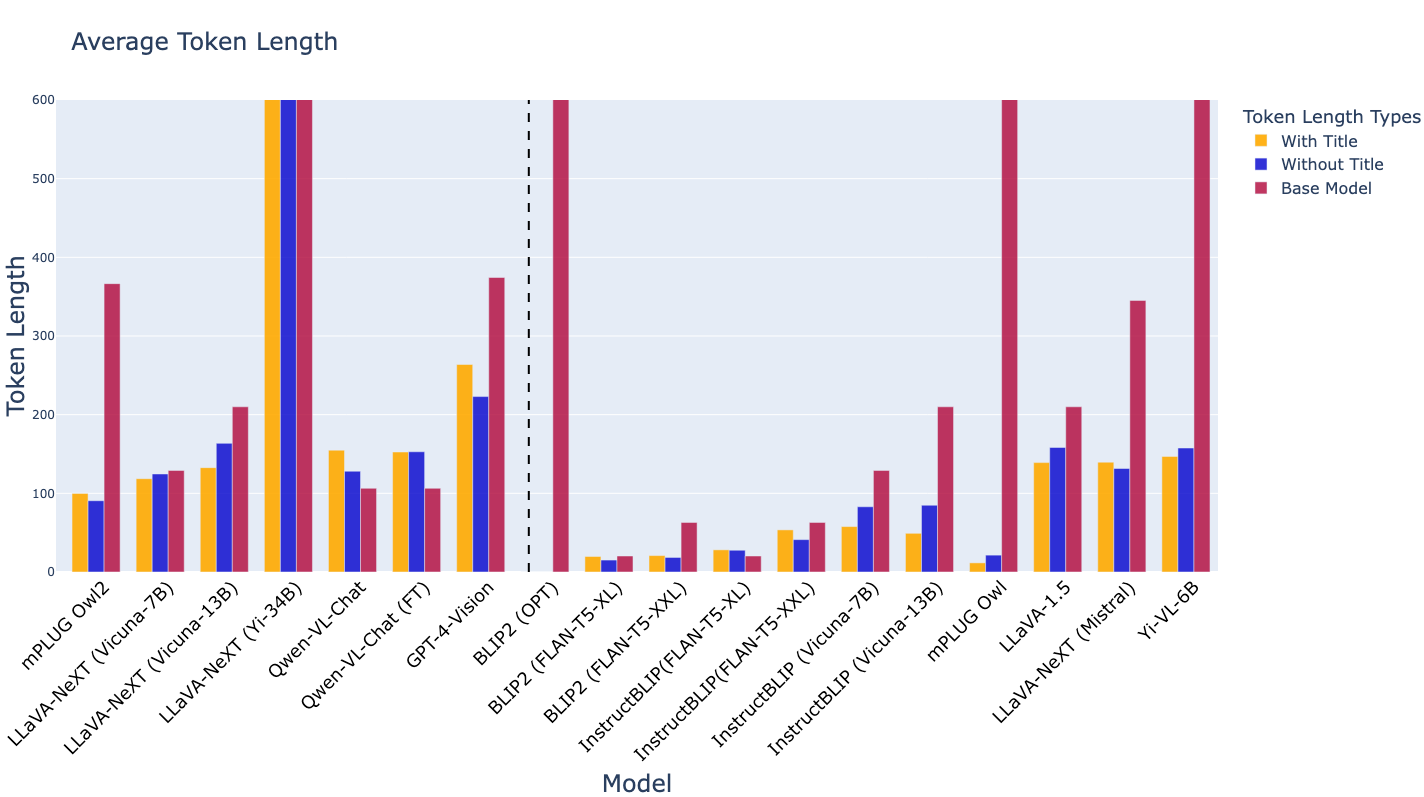}
\caption{Average token lengths for 18 evaluated LVLMs on an unseen set, where yellow represents the 'With Title' setting, bleu indicates the 'Without Title' setting, and red signifies the average token length for the base language model of the LVLM with titles. The length of the unseen reference sentence is 174 tokens.}
\label{fig:Token length}
\end{figure*}

%% file: figure/fig3.5.tex
\begin{figure*}[ht]
\centering
\includegraphics[width=\textwidth]{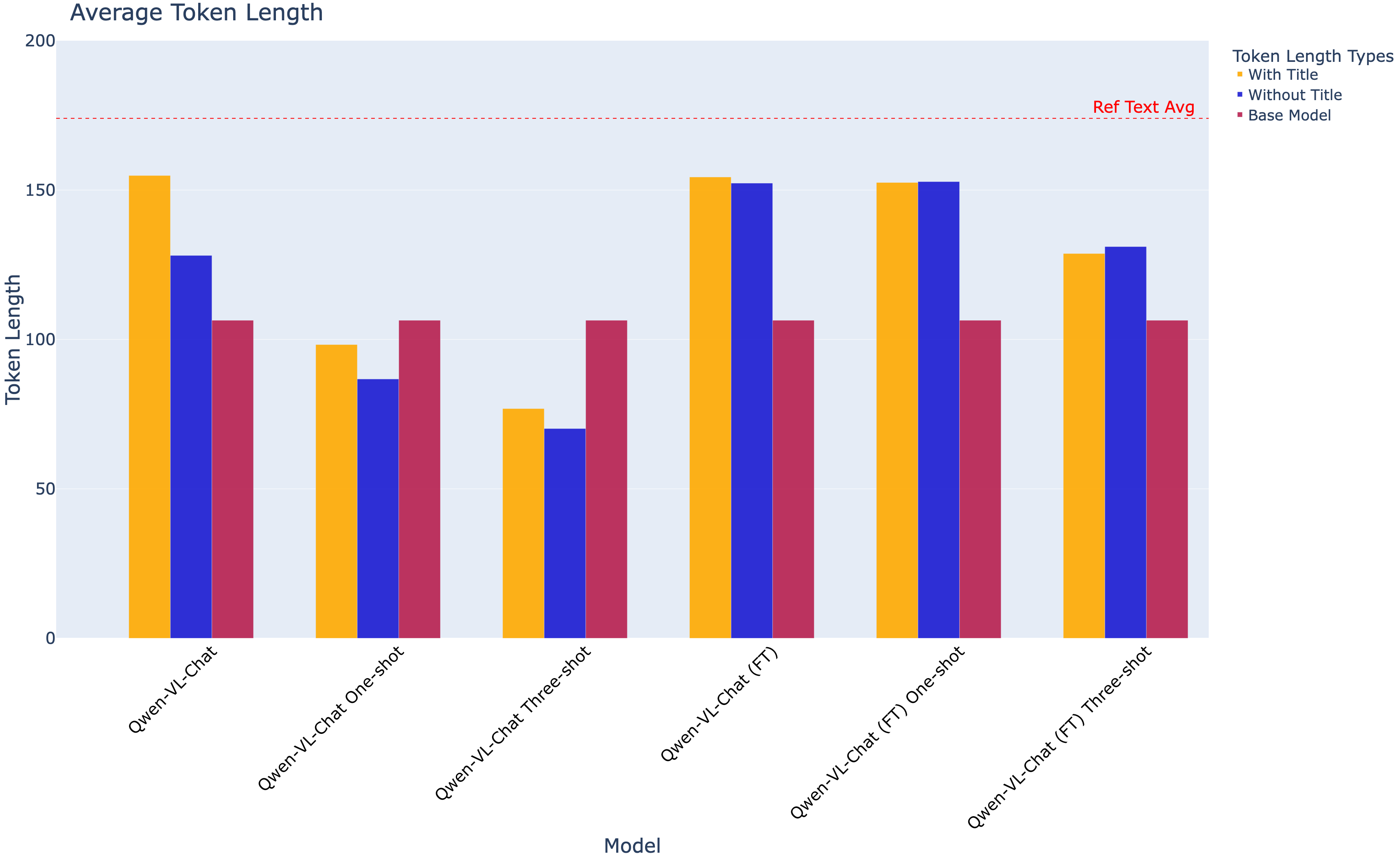}
\caption{Average token lengths for Qwen's Few-shot and Fine-tuning settings on an unseen set, where yellow represents the 'With Title' setting, bleu indicates the 'Without Title' setting, and red signifies the average token length for the base language model of the LVLM with titles. The length of the unseen reference sentence is 174 tokens.}
\label{fig:Qwen Token length}
\end{figure*}

%% file: table/table_6.tex
\begin{table*}[ht]
\centering
\resizebox{1.0\textwidth}{!}{
\footnotesize
\setlength{\tabcolsep}{6pt}
\begin{tabular}{@{}lccccccccccccccc@{}}
\toprule
\multicolumn{1}{c}{\multirow{2}{*}{LVLM}} & \multirow{2}{*}{Setting} & \multirow{2}{*}{Size} & \multirow{2}{*}{BLEU} & \multicolumn{3}{c}{ROUGE} & \multirow{2}{*}{BertScore} & \multicolumn{2}{c}{Entity Cov.} & \multirow{2}{*}{Entity F1} & \multicolumn{4}{c}{Entity Cooccurrence} & \multirow{2}{*}{Avg. Length} \\
\cmidrule(lr){5-7} \cmidrule(lr){9-10} \cmidrule(lr){12-15}
& & & & 1 & 2 & L & & exact & partial & & n=0 & n=1 & n=2 & n=$\infty$ & \\
\midrule
\rowcolor[rgb]{0.956, 0.965, 0.592}
\multicolumn{16}{c}{\textbf{\textit{With Title} (Language information + Visual information)}} \\
\midrule
FLAN-T5-XL & Unseen & 3B & 0.66 & 15.4 & 6.23 & 13.1 & 83.6 & 10.2 & 15.4 & 10.6 & 1.36 & 0.88 & 0.84 & 0.83 & 20\\
FLAN-T5-XXL & Unseen & 11B & 0.00 & 2.0 & 0.09 & 1.8 & 76.2 & 3.3 & 2.2 & 0.3 & 0.00 & 0.00 & 0.00 & 0.00 & 63 \\
OPT & Unseen & 6.7B & 0.34 & 8.3 & 1.60 & 7.3 & 76.8 & 12.0 & 18.9 & 8.4 & 0.15 & 0.12 & 0.12 & 0.11 & 872\\
LlaMA & Unseen & 7B & 0.48 & 9.4 & 1.99 & 8.1 & 77.7 & 16.4 & 23.7 & 11.3 &  0.15 & 0.14 & 0.13 & 0.11 & 876\\
LLaMA2 & Unseen & 7B & 1.81 & 24.0 & 5.92 & 14.9 & 82.4 & 18.5 & 27.3 & 20.8 & 1.04 & 0.88 & 0.82 & 0.81 & 366\\
Mistral & Unseen & 7B & 1.82 & 25.1 & 6.41 & 15.2 & 82.7 & 21.8 & 31.2 & 23.4 & 1.33 & 1.30 & 1.27 & 1.25 & 345\\
Vicuna-7B  & Unseen & 7B & 1.14 & 20.9 & 4.87 & 13.1 & 82.7 & 12.3 & 18.6 & 14.1 & 1.43 & 1.33 & 1.32 & 1.23 & 129\\
Vicuna-13B & Unseen & 13B & 2.35 & 28.4 & 7.34 & 17.7 & 83.4 & 19.4 & 28.1 & 23.0 & 2.16 & 1.99 & 1.89 & 1.77 & 210\\
Qwen-Chat & Unseen & 7B & 0.60 & 12.0 & 2.50 & 7.4 & 79.5 & 7.6 & 11.8 & 8.5 & 0.52 & 0.43 & 0.41 & 0.40 & 106 \\
Yi-6B-Chat & Unseen & 6B & 0.93 & 14.0 & 3.55 & 10.9 & 79.3 & 14.2 & 21.4 & 11.9 & 0.55 & 0.50 & 0.48 & 0.46 & 717\\
Yi-34B-Chat & Unseen & 34B & 1.00 & 13.1 & 3.50 & 10.4 & 79.1 & 17.9 & 25.4 & 12.9 & 0.93 & 0.86 & 0.83 & 0.81 & 745\\
GPT-4 & Unseen & - & 2.20 & 26.2 & 7.00 & 14.9 & 82.5 & 31.7 & 40.2 & 32.3 & 2.54 & 2.50 & 2.53 & 2.59  & 374\\
\midrule
FLAN-T5-XL & Seen & 3B & 0.67 & 15.1 & 6.30 & 12.9 & 83.4 & 9.0 & 14.5 & 9.5 & 1.34 & 0.95 & 0.85 & 0.81& 22\\
FLAN-T5-XXL & Seen & 11B & 0.01 & 8.9 & 1.48 & 7.5 & 81.2 & 2.1 & 5.0 & 1.1 & 0.01 & 0.00 & 0.00 & 0.00 & 66\\
OPT & Seen & 6.7B & 0.35 & 8.3 & 1.63 & 7.2 & 76.8 & 11.4 & 18.4 & 9.0 & 0.08 & 0.06 & 0.05 & 0.05 & 877\\
LlaMA & Seen & 7B & 0.51 & 9.3 & 2.01 & 8.0 & 77.8 & 15.7 & 23.1 & 11.0 & 0.17 & 0.13 & 0.12 & 0.10 & 877\\
LLaMA2 & Seen & 7B & 1.87 & 24.3 & 6.03 & 15.1 & 82.5 & 19.0 & 28.1 & 21.4 & 1.10 & 0.92 & 0.85 & 0.84 & 357 \\
Mistral & Seen & 7B & 1.91 & 25.1 & 6.40 & 15.2 & 82.6 & 20.3 & 29.5 & 22.5 & 1.33 & 1.11 & 1.03 & 0.98 & 334\\
Vicuna-7B  & Seen & 7B & 0.98 & 19.6 & 4.42 & 12.3 & 82.6 & 10.0 & 15.9 & 11.8 & 1.03 & 0.92 & 0.86 & 0.83 & 111\\
Vicuna-13B & Seen & 13B & 1.91 & 25.1 & 6.37 & 15.2 & 82.6 & 20.3 & 29.5 & 22.5 & 1.33 & 1.11 & 1.03 & 0.98 & 334\\
Qwen-Chat & Seen & 7B & 0.62 & 11.9 & 2.47 & 7.3 & 79.4 & 7.4 & 11.7 & 8.3 & 0.64 & 0.52 & 0.51 & 0.48 & 104\\
Yi-6B-Chat & Seen & 6B & 0.99 & 14.6 & 3.74 & 11.2 & 79.6 & 13.9 & 21.3 & 12.6 & 0.64 & 0.60 & 0.57 & 0.55 & 698\\
Yi-34B-Chat & Seen & 34B & 1.00 & 12.9 & 3.41 & 10.3 & 79.0 & 17.6 & 24.8 & 12.7 & 0.92 & 0.85 & 0.81 & 0.79 & 750\\
GPT-4 & Seen & - &2.20 & 26.0 & 6.90 & 14.8 & 82.5 & 29.7 & 38.3 & 31.0 & 2.50 & 2.30 & 2.32 & 2.31 & 369\\

\bottomrule
\end{tabular}
}
\caption{Comprehensive Performance of Base Language Models with title Integration. This table showcases the performance of primary models, both featured and not featured in the main analysis, across 'seen' and 'unseen' settings, evaluated using additional metrics such as BLEU, BERTscore, and ROUGE.}
\label{tab:Appendix LLM result with title setting}
\end{table*}

%% file: table/table_7.tex
\begin{table*}
    \centering
    \resizebox{\textwidth}{!}{
    \begin{tabular}{lcccccccc}
    \toprule
         & mPlug\_owl2 & LlaVA-NeXT (Vicuna13B) & LlaVA-NeXT (Vicuna7B) &LLaVA-NeXT (Yi34B)& Qwen-VL-Chat & Qwen-VL-Chat (FT) &  GPT-4-Vision \\ 
        \midrule
    Exact match & 1.6\% & 0.0\% & 0.0\% & 0.0\% & 4.0\% & 5.7\% & \textbf{8.97\%}  \\
    Partial match& 54.2\% & 39.9\% & 27.5\% & 66.3\% & 53.6\% & \textbf{66.7\%} & 64.0\%   \\
        \bottomrule
    \end{tabular}
    }
    \caption{LVLM Primary Group Analysis of Title Generation Accuracy from Image Information.}
    \label{tab:Appendix Main Model title generation}
\end{table*}

%% file: table/table_8.tex
\begin{table*}
    \centering
    \resizebox{\textwidth}{!}{
    \begin{tabular}{lcccccc}
    \toprule
         Setting & BLIP2 (OPT) & BLIP2 (FLAN-T5-XL) & BLIP2 (FLAN-T5-XXL) & mPLUG\_Owl & LLaVA-1.5 & InstructBLIP (FLAN-T5-XL) \\ 
        \midrule
    Exact match & 0.0\% & 1.04\% & 1.25\% & 1.97\% & 0.0\% & 0.93\% \\
    Partial match & 0.10\% & 49.6\% & 49.1\% & 37.0\% & 40.3\% & 44.0\% \\
        \bottomrule
    \end{tabular}
    }
    \caption{LVLM Complementary Group Analysis of Title Generation Accuracy Using Only Image Information (Part 1).}
    \label{tab:Appendix Sub Model title generation part1}
\end{table*}

\begin{table*}
    \centering
    \resizebox{\textwidth}{!}{
    \begin{tabular}{lcccccccc}
    \toprule
         Setting & InstructBLIP (FLAN-T5-XXL) & InstructBLIP (Vicuna-7B) & Instruct Blip (Vicuna-13B) & LLaVA-NeXT (mistral) & Yi-VL-6B \\ 
        \midrule
    Exact match & 1.04\% & 1.14\% & 1.14\% & 0.10\% & 1.36\% \\
    Partial match & 50.1\% & 50.5\% & 58.1\% & 47.7\% & 50.6\% \\
        \bottomrule
    \end{tabular}
    }
    \caption{LVLM Complementary Group Analysis of Title Generation Accuracy Using Only Image Information (Part 2).}
    \label{tab:Appendix Sub Model title generation part2}
\end{table*}

%% file: table/table_13.tex
\begin{table*}[ht]
    \scriptsize
    \centering
    \begin{tabularx}{\textwidth}{lXX}
        \toprule
        \textbf{Type} & \textbf{Title-Included Template} & \textbf{Title-Excluded Template} \\
        \midrule
        Section &
        \texttt{Explain the \textcolor{red}{\{Section\}} of this artwork, \textbf{\textcolor{blue}{\{Title\}}}.} &
        \texttt{Explain the \textcolor{red}{\{Section\}} of this artwork.} \\
        Subsection &
        \texttt{Explain the \textcolor{red}{\{Subsection\}} about the \textcolor{red}{\{Section\}} of this artwork, \textbf{\textcolor{blue}{\{Title\}}}.} &
        \texttt{Explain the \textcolor{red}{\{Subsection\}} about the \textcolor{red}{\{Section\}} of this artwork.} \\
        Sub subsection &
        \texttt{Explain the \textcolor{red}{\{Sub subsection\}} about the \textcolor{red}{\{Subsection\}} of the \textcolor{red}{\{Section\}} in this artwork, \textbf{\textcolor{blue}{\{Title\}}}.} &
        \texttt{Explain the \textcolor{red}{\{Sub subsection\}} about the \textcolor{red}{\{Subsection\}} of the \textcolor{red}{\{Section\}} in this artwork.} \\
        \bottomrule
    \end{tabularx}
    \caption{Prompt templates used in the Test split.  
    We employ a single controlled template with hierarchical granularity (Section, Subsection, Sub subsection),  
    each with Title-Included and Title-Excluded variants.}
    \label{tab:Prompt_Test_Templates}
\end{table*}

\begin{table*}[ht]
    \scriptsize
    \centering
    \begin{tabularx}{\textwidth}{lXX}
        \toprule
        \textbf{Type} & \textbf{Title-Included Template} & \textbf{Title-Excluded Template} \\
        \midrule
        Section &
        \texttt{Describe the \textcolor{red}{\{Section\}} of this artwork, \textbf{\textcolor{blue}{\{Title\}}}.} &
        \texttt{Describe the \textcolor{red}{\{Section\}} of this artwork.} \\
        Subsection &
        \texttt{Describe the \textcolor{red}{\{Subsection\}} about the \textcolor{red}{\{Section\}} of this artwork, \textbf{\textcolor{blue}{\{Title\}}}.} &
        \texttt{Describe the \textcolor{red}{\{Subsection\}} about the \textcolor{red}{\{Section\}} of this artwork.} \\
        Sub subsection &
        \texttt{Describe the \textcolor{red}{\{Sub subsection\}} about the \textcolor{red}{\{Subsection\}} of the \textcolor{red}{\{Section\}} in this artwork, \textbf{\textcolor{blue}{\{Title\}}}.} &
        \texttt{Describe the \textcolor{red}{\{Sub subsection\}} about the \textcolor{red}{\{Subsection\}} of the \textcolor{red}{\{Section\}} in this artwork.} \\
        \bottomrule
    \end{tabularx}
    \caption{Prompt templates used in the Dev split.  
    We employ a single controlled template with hierarchical granularity (Section, Subsection, Sub subsection),  
    each with Title-Included and Title-Excluded variants.}
    \label{tab:Prompt_Dev_Templates}
\end{table*}

\begin{table*}[ht]
    \scriptsize
    \centering
    \setlength{\tabcolsep}{4pt}
    \renewcommand{\arraystretch}{1.05}
    \begin{tabularx}{\textwidth}{l>{\centering\arraybackslash}p{2.2cm}XX}
        \toprule
        \textbf{Template} & \textbf{Type} & \textbf{Title-Included Template} & \textbf{Title-Excluded Template} \\
        \midrule

        \multirow{3}{*}{\textbf{Template 1}}
        & Section
        & \texttt{Focus on \textbf{\textcolor{blue}{\{Title\}}} and explore the \textcolor{red}{\{Section\}}.}
        & \texttt{Focus on this artwork and explore the \textcolor{red}{\{Section\}}.} \\
        & Subsection
        & \texttt{In the context of \textbf{\textcolor{blue}{\{Title\}}}, explore the \textcolor{red}{\{Subsection\}} of the \textcolor{red}{\{Section\}}.}
        & \texttt{In the context of this artwork, explore the \textcolor{red}{\{Subsection\}} of the \textcolor{red}{\{Section\}}.} \\
        & Sub subsection
        & \texttt{Focusing on the \textcolor{red}{\{Section\}} of \textbf{\textcolor{blue}{\{Title\}}}, explore the \textcolor{red}{\{Sub subsection\}} about the \textcolor{red}{\{Subsection\}}.}
        & \texttt{Focusing on the \textcolor{red}{\{Section\}} of this artwork, explore the \textcolor{red}{\{Sub subsection\}} about the \textcolor{red}{\{Subsection\}}.} \\
        \midrule

        \multirow{3}{*}{\textbf{Template 2}}
        & Section
        & \texttt{Focus on \textbf{\textcolor{blue}{\{Title\}}} and explain the \textcolor{red}{\{Section\}}.}
        & \texttt{Focus on this artwork and explain the \textcolor{red}{\{Section\}}.} \\
        & Subsection
        & \texttt{In the context of \textbf{\textcolor{blue}{\{Title\}}}, explain the \textcolor{red}{\{Subsection\}} of the \textcolor{red}{\{Section\}}.}
        & \texttt{In the context of this artwork, explain the \textcolor{red}{\{Subsection\}} of the \textcolor{red}{\{Section\}}.} \\
        & Sub subsection
        & \texttt{Focusing on the \textcolor{red}{\{Section\}} of \textbf{\textcolor{blue}{\{Title\}}}, explain the \textcolor{red}{\{Sub subsection\}} about the \textcolor{red}{\{Subsection\}}.}
        & \texttt{Focusing on the \textcolor{red}{\{Section\}} of this artwork, explain the \textcolor{red}{\{Sub subsection\}} about the \textcolor{red}{\{Subsection\}}.} \\
        \midrule

        \multirow{3}{*}{\textbf{Template 3}}
        & Section
        & \texttt{Explore the \textcolor{red}{\{Section\}} of this artwork, \textbf{\textcolor{blue}{\{Title\}}}.}
        & \texttt{Explore the \textcolor{red}{\{Section\}} of this artwork.} \\
        & Subsection
        & \texttt{Explore the \textcolor{red}{\{Subsection\}} about the \textcolor{red}{\{Section\}} of this artwork, \textbf{\textcolor{blue}{\{Title\}}}.}
        & \texttt{Explore the \textcolor{red}{\{Subsection\}} about the \textcolor{red}{\{Section\}} of this artwork.} \\
        & Sub subsection
        & \texttt{Explore the \textcolor{red}{\{Sub subsection\}} about the \textcolor{red}{\{Subsection\}} of the \textcolor{red}{\{Section\}} in this artwork, \textbf{\textcolor{blue}{\{Title\}}}.}
        & \texttt{Explore the \textcolor{red}{\{Sub subsection\}} about the \textcolor{red}{\{Subsection\}} of the \textcolor{red}{\{Section\}} in this artwork.} \\
        \midrule

        \multirow{3}{*}{\textbf{Template 4}}
        & Section
        & \texttt{Focus on \textbf{\textcolor{blue}{\{Title\}}} and discuss the \textcolor{red}{\{Section\}}.}
        & \texttt{Focus on this artwork and discuss the \textcolor{red}{\{Section\}}.} \\
        & Subsection
        & \texttt{In the context of \textbf{\textcolor{blue}{\{Title\}}}, discuss the \textcolor{red}{\{Subsection\}} of the \textcolor{red}{\{Section\}}.}
        & \texttt{In the context of this artwork, discuss the \textcolor{red}{\{Subsection\}} of the \textcolor{red}{\{Section\}}.} \\
        & Sub subsection
        & \texttt{Focusing on the \textcolor{red}{\{Section\}} of \textbf{\textcolor{blue}{\{Title\}}}, discuss the \textcolor{red}{\{Sub subsection\}} about the \textcolor{red}{\{Subsection\}}.}
        & \texttt{Focusing on the \textcolor{red}{\{Section\}} of this artwork, discuss the \textcolor{red}{\{Sub subsection\}} about the \textcolor{red}{\{Subsection\}}.} \\
        \midrule

        \multirow{3}{*}{\textbf{Template 5}}
        & Section
        & \texttt{How does \textbf{\textcolor{blue}{\{Title\}}} elucidate its \textcolor{red}{\{Section\}}?}
        & \texttt{How does this artwork elucidate its \textcolor{red}{\{Section\}}?} \\
        & Subsection
        & \texttt{In \textbf{\textcolor{blue}{\{Title\}}}, how is the \textcolor{red}{\{Subsection\}} of the \textcolor{red}{\{Section\}} elucidated?}
        & \texttt{In this artwork, how is the \textcolor{red}{\{Subsection\}} of the \textcolor{red}{\{Section\}} elucidated?} \\
        & Sub subsection
        & \texttt{Regarding \textbf{\textcolor{blue}{\{Title\}}}, how does the \textcolor{red}{\{Section\}}'s \textcolor{red}{\{Subsection\}} incorporate the \textcolor{red}{\{Sub subsection\}}?}
        & \texttt{Regarding this artwork, how does the \textcolor{red}{\{Section\}}'s \textcolor{red}{\{Subsection\}} incorporate the \textcolor{red}{\{Sub subsection\}}?} \\
        \midrule

        \multirow{3}{*}{\textbf{Template 6}}
        & Section
        & \texttt{Focus on \textbf{\textcolor{blue}{\{Title\}}} and analyze the \textcolor{red}{\{Section\}}.}
        & \texttt{Focus on this artwork and analyze the \textcolor{red}{\{Section\}}.} \\
        & Subsection
        & \texttt{In the context of \textbf{\textcolor{blue}{\{Title\}}}, analyze the \textcolor{red}{\{Subsection\}} of the \textcolor{red}{\{Section\}}.}
        & \texttt{In the context of this artwork, analyze the \textcolor{red}{\{Subsection\}} of the \textcolor{red}{\{Section\}}.} \\
        & Sub subsection
        & \texttt{Focusing on the \textcolor{red}{\{Section\}} of \textbf{\textcolor{blue}{\{Title\}}}, analyze the \textcolor{red}{\{Sub subsection\}} about the \textcolor{red}{\{Subsection\}}.}
        & \texttt{Focusing on the \textcolor{red}{\{Section\}} of this artwork, analyze the \textcolor{red}{\{Sub subsection\}} about the \textcolor{red}{\{Subsection\}}.} \\
        \midrule

        \multirow{3}{*}{\textbf{Template 7}}
        & Section
        & \texttt{In \textbf{\textcolor{blue}{\{Title\}}}, how is the \textcolor{red}{\{Section\}} discussed?}
        & \texttt{In this artwork, how is the \textcolor{red}{\{Section\}} discussed?} \\
        & Subsection
        & \texttt{Describe the characteristics of the \textcolor{red}{\{Subsection\}} in \textbf{\textcolor{blue}{\{Title\}}}'s \textcolor{red}{\{Section\}}.}
        & \texttt{Describe the characteristics of the \textcolor{red}{\{Subsection\}} in this artwork's \textcolor{red}{\{Section\}}.} \\
        & Sub subsection
        & \texttt{When looking at the \textcolor{red}{\{Section\}} of \textbf{\textcolor{blue}{\{Title\}}}, how do you discuss its \textcolor{red}{\{Subsection\}}'s \textcolor{red}{\{Sub subsection\}}?}
        & \texttt{When looking at the \textcolor{red}{\{Section\}} of this artwork, how do you discuss its \textcolor{red}{\{Subsection\}}'s \textcolor{red}{\{Sub subsection\}}?} \\
        \bottomrule
    \end{tabularx}
    \caption{Prompt templates used in the Train split.
    Each template is instantiated at three hierarchical levels (Section, Subsection, Sub subsection),
    with both Title-Included and Title-Excluded variants to diversify linguistic realizations
    while preserving the underlying structural format.}
    \label{tab:Prompt_Train_Templates}
\end{table*}

\begin{table*}[t]
\centering
\small
\begin{tabular}{l l r r}
\toprule
\textbf{Split} & \textbf{Category} & \textbf{Count} & \textbf{Ratio (\%)} \\
\midrule
\multirow{7}{*}{\textbf{Dev}} 
 & Title-Included Section & 2434 & 44.06 \\
 & Title-Included Subsection & 306 & 5.54 \\
 & Title-Included Sub subsection & 22 & 0.40 \\
 & Title-Excluded Section & 2434 & 44.06 \\
 & Title-Excluded Subsection & 306 & 5.54 \\
 & Title-Excluded Sub subsection & 22 & 0.40 \\
\cmidrule{2-4}
 & \textbf{Total} & \textbf{5524} & \textbf{100.00} \\
\midrule
\multirow{7}{*}{\textbf{Test}} 
 & Title-Included Section & 4649 & 44.47 \\
 & Title-Included Subsection & 538 & 5.15 \\
 & Title-Included Sub subsection & 40 & 0.38 \\
 & Title-Excluded Section & 4649 & 44.47 \\
 & Title-Excluded Subsection & 538 & 5.15 \\
 & Title-Excluded Sub subsection & 40 & 0.38 \\
\cmidrule{2-4}
 & \textbf{Total} & \textbf{10454} & \textbf{100.00} \\
\midrule
\multirow{11}{*}{\textbf{Train}} 
 & Template 1 & 5476 & 14.27 \\
 & Template 2 & 5496 & 14.33 \\
 & Template 3 & 5428 & 14.15 \\
 & Template 4 & 5502 & 14.34 \\
 & Template 5 & 5460 & 14.23 \\
 & Template 6 & 5502 & 14.34 \\
 & Template 7 & 5498 & 14.33 \\
\cmidrule{2-4}
 & Section (all templates) & 33582 & 87.54 \\
 & Subsection (all templates) & 4326 & 11.28 \\
 & Sub subsection (all templates) & 454 & 1.18 \\
\cmidrule{2-4}
 & \textbf{Total} & \textbf{38362} & \textbf{100.00} \\
\bottomrule
\end{tabular}
\caption{Template statistics for the Train, Dev, and Test splits.
Dev and Test employ a single controlled template supporting hierarchical granularity
(Section, Subsection, Sub subsection) with Title-Included/Excluded variants.
Train follows the same structural template format but introduces seven linguistic variants
to avoid overfitting to a particular prompt style, and we additionally report the distribution of hierarchy levels in Train.}
\label{tab:dataset_stats}
\end{table*}

%% file: supplement/with_title.tex
\begin{figure}[ht]
\centering
\begin{tabular}{c}
{\scriptsize
\begin{lstlisting}[language=json, breakatwhitespace=false]
{
  "id": "0001_T",
  "title": "Mona Lisa",
  "conversations": [
    {
      "from": "user",
      "value": "<img>/images/Mona Lisa.jpg</img>\nFocus on Mona Lisa and explore the history."
    },
    {
      "from": "assistant",
      "value": "Of Leonardo da Vinci's works, the Mona Lisa is the only portrait whose authenticity...."
    }
  ]
}
\end{lstlisting}
}
\end{tabular}
\caption{Train set format with title.}
\label{tab:train_set_format_with_title}
\end{figure}

%% file: supplement/without_title.tex
\begin{figure}[ht]
\centering
\begin{tabular}{c}
{\scriptsize
\begin{lstlisting}[language=json, breakatwhitespace=false]
{
  "id": "0001_NT",
  "conversations": [
    {
      "from": "user",
      "value": "<img>/images/Mona Lisa.jpg</img>\nFocus on this artwork and explore the history."
    },
    {
      "from": "assistant",
      "value": "Of Leonardo da Vinci's works, the Mona Lisa is the only portrait whose authenticity...."
    }
  ]
}
\end{lstlisting}
}
\end{tabular}
\caption{Train set format without title.}
\label{tab:train_set_format_without_title}
\end{figure}

%% file: figure/fig4.tex
\begin{figure*}[ht]
\centering
\includegraphics[width=\textwidth]{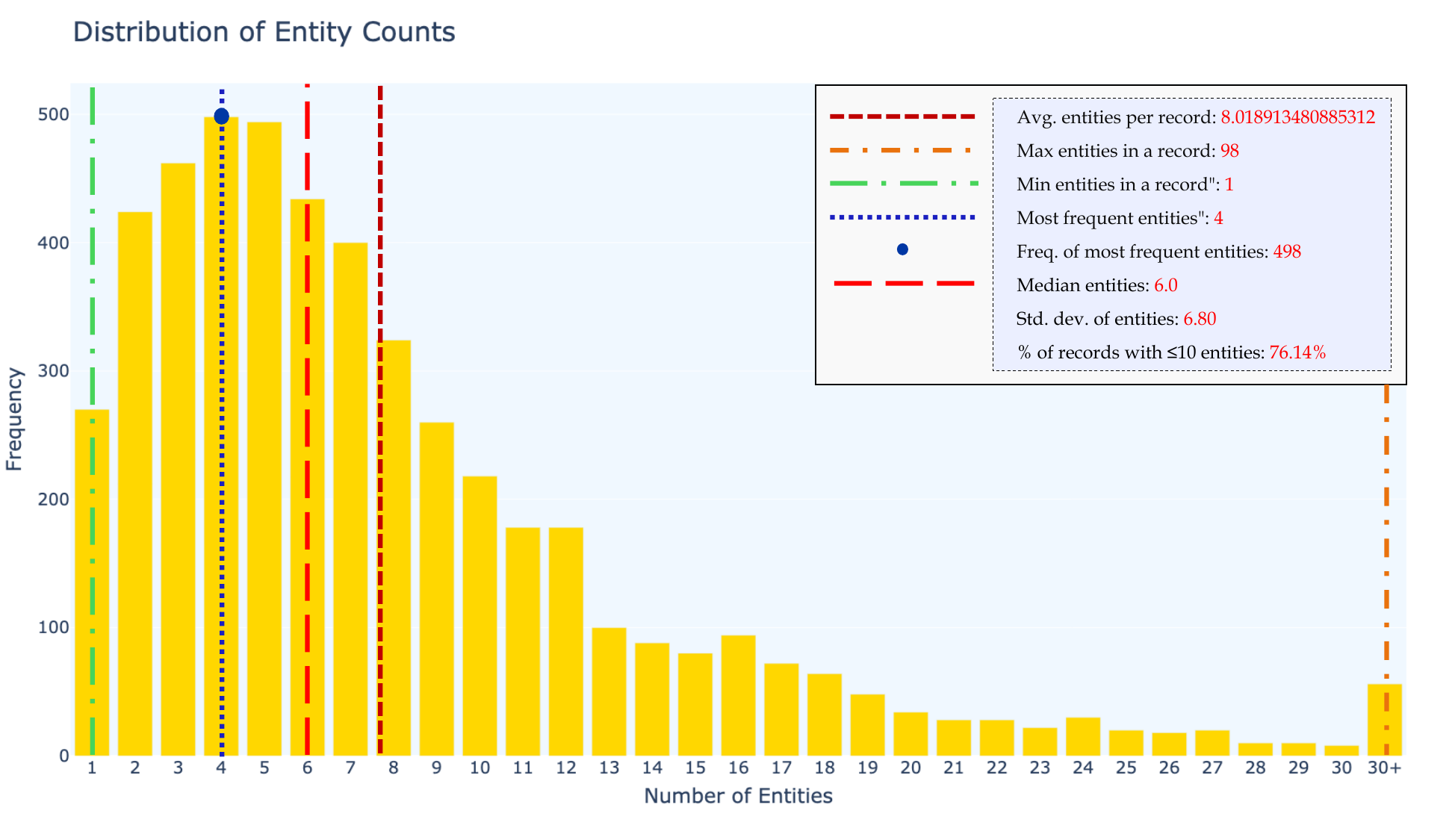}
\caption{Entity distribution within each dataset under the 'With Title' setting.}
\label{fig:title_entity}
\end{figure*}

%% file: figure/fig5.tex
\begin{figure*}[ht]
\centering
\includegraphics[width=\textwidth]{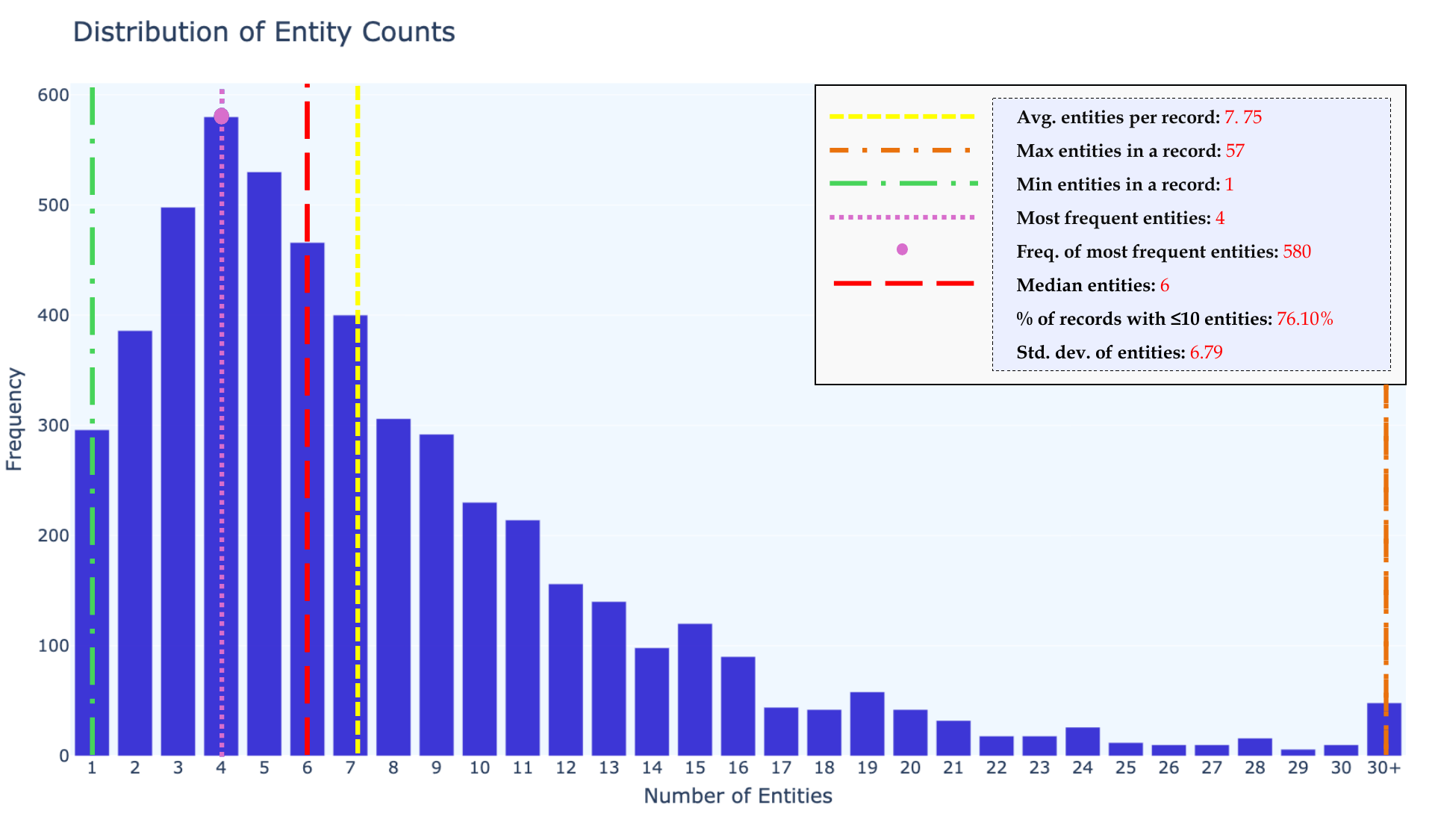}
\caption{Entity distribution within each dataset under the 'without title' setting.}
\label{fig:nontitle_entity}
\end{figure*}

%% file: supplement/mytable.tex
\begin{table}[ht]
\centering
\scriptsize
\begin{tabular}{llcccc}
    \toprule
    \textbf{Data Type} & \textbf{Data Name} & \textbf{mPlug-owl} & \textbf{Qwen-VL-Chat} & \textbf{LLava-v-1.5} & \textbf{InstructBLIP} \\
    \midrule
    Text & ShareGPT \cite{chen2023sharegpt4v} & \CheckmarkBold &  & \CheckmarkBold &  \\
    & SlimOrca \cite{mukherjee2023orca} & \CheckmarkBold &  &  &  \\
    & In-house Data &  & \CheckmarkBold &  &  \\
    Dialogue & LLaVA \cite{liu2023llava} & \CheckmarkBold &  & \CheckmarkBold &  \\
    Caption & COCO \cite{Lin2014MicrosoftCC} & \CheckmarkBold & \CheckmarkBold &  & \CheckmarkBold \\
    & TextCaps \cite{sidorov2020textcaps}& \CheckmarkBold &  & \CheckmarkBold & \CheckmarkBold \\
    & SBU \cite{NIPS2011_5dd9db5e} &  & \CheckmarkBold &  &  \\
    & Coyo \cite{kakaobrain2022coyo-700m} &  & \CheckmarkBold &  &  \\
    & DataComp  &  & \CheckmarkBold &  &  \\
    & CC12M \& 3M \cite{changpinyo2021cc12m} &  & \CheckmarkBold &  &  \\
    & LAION-en \cite{schuhmann2022laion5b} \& zh &  & \CheckmarkBold &  &  \\
    VQA & VQAv2 & \CheckmarkBold & \CheckmarkBold & \CheckmarkBold & \CheckmarkBold \\
    & GQA \cite{8953451} & \CheckmarkBold & \CheckmarkBold & \CheckmarkBold & \CheckmarkBold \\
    & OKVQA \cite{8953725} & \CheckmarkBold &  & \CheckmarkBold & \CheckmarkBold \\
    & OCRVQA \cite{8978122} & \CheckmarkBold & \CheckmarkBold & \CheckmarkBold & \CheckmarkBold \\
    & A-OKVQA \cite{schwenk2022aokvqabenchmarkvisualquestion}& \CheckmarkBold &  & \CheckmarkBold & \CheckmarkBold \\
    & DVQA \cite{kafle2018dvqa} &  & \CheckmarkBold &  &  \\
    & TextVQA \cite{Singh2019TowardsVM}&  & \CheckmarkBold & \CheckmarkBold & \CheckmarkBold \\
    & ChartQA \cite{masry-etal-2022-chartqa} &  & \CheckmarkBold &  &  \\
    & A12D &  & \CheckmarkBold &  &  \\
    Grounding² & GRIT \cite{Kosmos2} &  & \CheckmarkBold &  &  \\
    Ref Grounding & GRIT &  & \CheckmarkBold &  &  \\
    & VisualGenome \cite{Krishna2016VisualGC} &  & \CheckmarkBold & \CheckmarkBold &  \\
    & RefCOCO \cite{yu2016modelingcontextreferringexpressions}&  & \CheckmarkBold & \CheckmarkBold &  \\
    & RefCOCO+ \cite{yu2016modelingcontextreferringexpressions}&  & \CheckmarkBold & \CheckmarkBold &  \\
    & RefCOCOg &  & \CheckmarkBold & \CheckmarkBold &  \\
    OCR & SynthDoG-en \cite{kim2022donut} \& zh &  & \CheckmarkBold &  &  \\
    & Common Crawl pdf \& HTML &  & \CheckmarkBold &  &  \\
    Image Captioning & Web CapFilt \cite{pmlr-v162-li22n}&  &  &  & \CheckmarkBold \\
    & NoCaps &  &  &  & \CheckmarkBold \\
    & Flickr30K \cite{7410660}&  &  &  & \CheckmarkBold \\
    Visual Spatial Reasoning & IconQA \cite{lu2021iconqa}&  &  &  & \CheckmarkBold \\
    Visual Dialog & Visual Dialog &  &  &  & \CheckmarkBold \\
    Video Question Answering & MSVD-QA \cite{xu2017video}&  &  &  & \CheckmarkBold \\
    & MSRVTT-QA &  &  &  & \CheckmarkBold \\
    & iVQA \cite{8578996}&  &  &  & \CheckmarkBold \\
    Image Classification & VizWiz \cite{Gurari2018VizWizGC}&  &  &  & \CheckmarkBold \\
    Knowledge-Grounded Image QA & ScienceQA \cite{lu2022learn}&  &  &  & \CheckmarkBold \\
    \bottomrule
\end{tabular}
\caption{Details of training datasets.}
\label{tab:training_datasets}
\end{table}